\documentclass[journal]{IEEEtran}
\usepackage{amsmath,amsfonts}
\usepackage{algorithmic}
\usepackage{algorithm}
\usepackage{array}
\usepackage[caption=false,font=normalsize,labelfont=sf,textfont=sf]{subfig}
\usepackage{textcomp}
\usepackage{stfloats}
\usepackage{url}
\usepackage{verbatim}
\usepackage{graphicx}
\usepackage{tcolorbox}
\usepackage{cite}
\usepackage{booktabs}
\usepackage{threeparttable}
\usepackage{multirow}

\usepackage{xcolor}
\usepackage[normalem]{ulem}
\newif\ifrevmarkup
\ifrevmarkup
  \newcommand{\added}[1]{\textcolor{blue}{#1}}
  \newcommand{\deleted}[1]{\textcolor{red}{\sout{#1}}}
  \newcommand{\replaced}[2]{\deleted{#2}\added{#1}}
\else
  \newcommand{\added}[1]{#1}
  \newcommand{\deleted}[1]{}
  \newcommand{\replaced}[2]{#1}
\fi

\begin{document}

\title{Disentangling Visual and Factual Correctness in LVLMs' Visualization Literacy}

\author{Soohyun Lee, Jaeyoung Kim, Seokhyeon Park, Sihyeon Lee, Jiwon Song, 
Bohyoung Kim, Hyunjoo Song, and Jinwook Seo
    \thanks{Soohyun Lee, Seokhyeon Park, Sihyeon Lee, Jiwon Song and Jinwook Seo are with Seoul National University. 
    E-mail: \{shlee, shpark, sihyeon, jwsong\}@hcil.snu.ac.kr; jseo@snu.ac.kr}%
    \thanks{Jaeyoung Kim is with MADI Co., Ltd.
    E-mail: jykim@madidt.com}%
    \thanks{Bohyoung Kim is with Hankuk University of Foreign Studies.
    E-mail: bkim@hufs.ac.kr}%
    \thanks{Hyunjoo Song is with Soongsil University.
    E-mail: hsong@ssu.ac.kr
    }
    \thanks{Soohyun Lee and Jaeyoung Kim contributed equally to this work (co-first authors).}%
    \thanks{Hyunjoo Song and Jinwook Seo are the corresponding authors.}
}

\markboth{Journal of \LaTeX\ Class Files,~Vol.~14, No.~8, August~2021}%
{Shell \MakeLowercase{\textit{et al.}}: A Sample Article Using IEEEtran.cls for IEEE Journals}


\maketitle

\begin{abstract}
Large Vision-Language Models (LVLMs) have shown remarkable capabilities in visualization interpretation, yet it remains unclear whether their responses reflect genuine reasoning over visual evidence or the influence of factual priors learned during training. Current evaluation methods mix these two sources, obscuring when correct visual interpretation is overridden by memorized factual knowledge.
We present a disentanglement framework that systematically isolates visual correctness from factual correctness, revealing fundamental validity limitations in existing visualization literacy assessments.
Through three complementary experiments with \replaced{15}{12} state-of-the-art LVLMs, we demonstrate that:
(1) Although several models achieve human-level performance on standard tests (VLAT), such performance may reflect factual recall rather than visual understanding, whereas randomized-data tests (reVLAT) underestimate visualization literacy when visual interpretation is correct but superseded by conflicting factual priors.
(2) Using our Counterfactual Visualization Literacy Assessment Test (CVLAT) \replaced{alongside capability-normalized arbitration metrics, we classify models by the sign of their visual--factual reliance index (VFRI). This classification reveals a visualization-oriented majority and a factual knowledge-oriented minority, although several near-zero cases warrant cautious interpretation. The factual knowledge-oriented minority tends to override the chart with}{, we classify models into two distinct groups: ten models consistently prioritize factual priors, whereas two models reliably follow visual encodings even when those encodings contradict} prior knowledge. \added{A human baseline ($N=30$) on the same counterfactual items confirms that people overwhelmingly follow the chart under conflict, providing a human reference point for visual--factual arbitration.}
(3) Prompt-based intervention can \replaced{shift this prioritization, but its effectiveness is highly model-dependent and often direction-asymmetric, with some models responding strongly to only one prompt direction. Furthermore, high chart-reading capability does not predict prompt-controllability}{partially shift this prioritization in about half of models, but the degree of controllability varies significantly by architecture}, indicating that visual-factual arbitration is not uniformly steerable.
Overall, our findings demonstrate that LVLMs’ high visualization accuracy is not sufficient evidence of faithful visual reasoning. We argue that reliable LVLM integration into visual analytics requires evaluating not only visualization literacy, but also how models arbitrate between visual evidence and factual priors, particularly when the two sources diverge. \added{The CVLAT benchmark and code are available at \url{https://github.com/JaeyoungKim-HCIL/CVLAT}.}
\end{abstract}


\section{Introduction}

\begin{figure}
\centering
\includegraphics[width=\columnwidth]{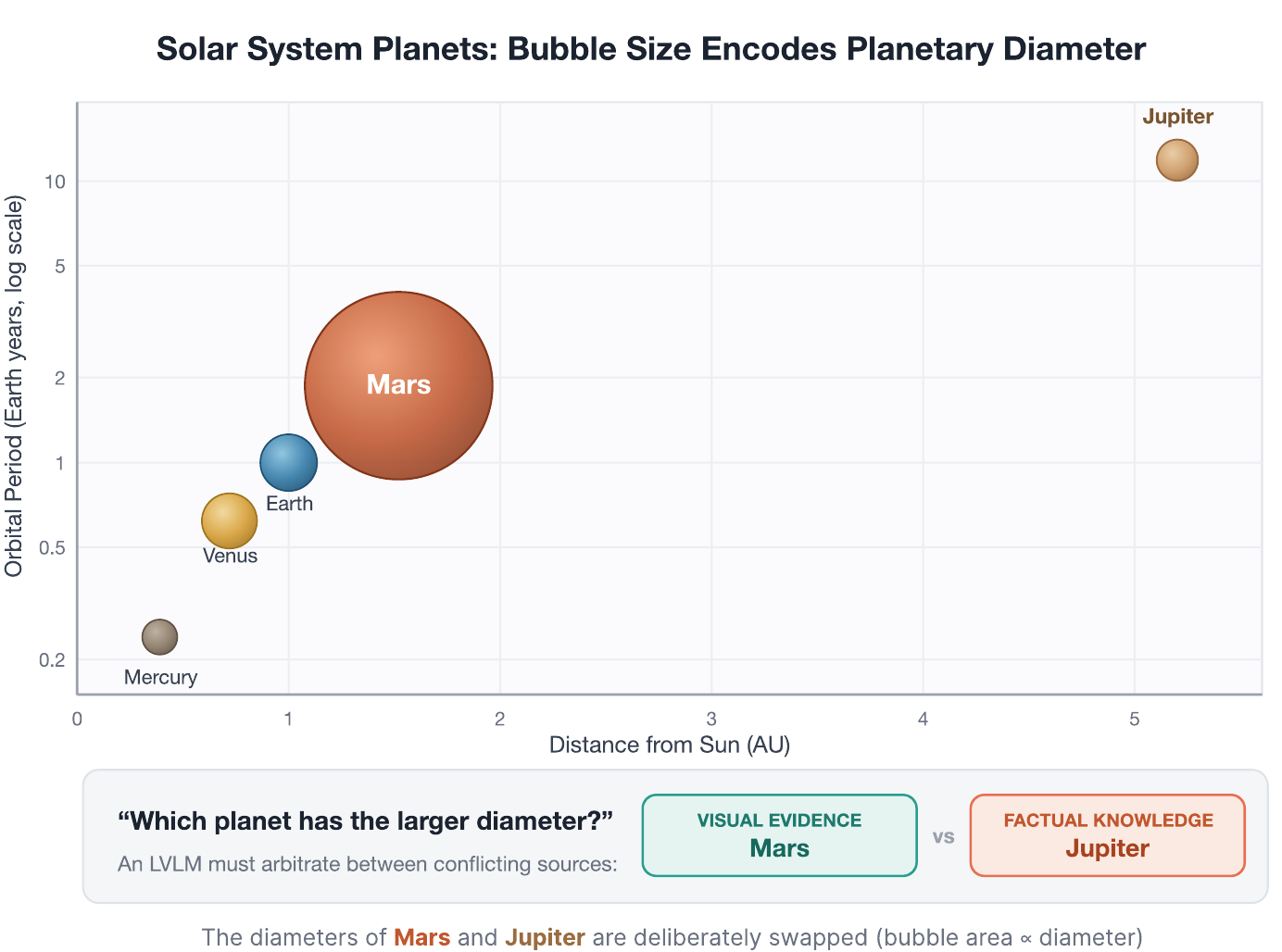}
\caption{A bubble chart of solar system planets where bubble size encodes planetary diameter, except Mars and Jupiter's sizes are deliberately swapped. When asked, ``Which planet has the larger diameter?" an LVLM must choose between visual encoding (Mars appears larger) and its pre-trained factual knowledge (Jupiter is larger). This conflict illustrates the core challenge in evaluating visualization literacy: distinguishing genuine visual interpretation from factual recall.
}
\label{fig:intro_counterfactual}
\end{figure}

Recent advances in Large Language Models (LLMs) and Large Vision-Language Models (LVLMs) have demonstrated strong capabilities across diverse visual analytics tasks, including data wrangling \cite{kim2024phenoflow, wang2023data, wang2025data}, visualization generation \cite{tian2024chartgpt, maddigan2023chat2vis, dibia2023lida, chen2025interchat}, and visualization interpretation \cite{jung2025can, wang2024scientific}.
Among these, visualization interpretation ability---commonly framed as visualization literacy---has received particular attention, as it plays a central role in determining whether LVLMs can reliably support analytical workflows, where accurate interpretation of charts and graphs is crucial for informed decision-making.
A growing body of work has empirically evaluated this ability \cite{bendeck2024empirical, hong2025llms, pandey2025benchmarking, li2024visualization, lo2024good, mukherjee2025encqa}, establishing valuable performance baselines while highlighting the limits of current evaluation practices.

Despite these foundational efforts, current evaluations leave two key questions unresolved. First, existing studies have primarily evaluated a limited set of proprietary LVLMs, such as GPT-4 \cite{achiam2023gpt}, Claude \added{3 Opus} \cite{anthropic2024claude3}, and Gemini \added{1.5 Pro} \cite{team2024gemini}, while rapid model development has introduced numerous newer and open-source alternatives that remain underexamined.
Second and more fundamentally, conventional accuracy-based assessment metrics fail to disentangle the basis of correctness---whether a correct answer reflects genuine visual interpretation or merely factual priors acquired during pretraining, and whether an incorrect answer stems from visual misreading or from overriding visual evidence with prior knowledge.

This second gap introduces a fundamental validity concern.
Consider the bubble chart in Figure~\ref{fig:intro_counterfactual}, where bubble size encodes planetary diameter, but the diameter values for Mars and Jupiter are intentionally reversed.
When asked \textit{``Which planet has the larger diameter?''}, an LVLM must choose between the visual evidence (Mars) and its pre-trained factual knowledge (Jupiter). If it responds ``Jupiter,'' existing accuracy metrics cannot determine whether it (1) correctly interpreted the chart but prioritized factual priors, or (2) failed to interpret the visualization in the first place. 
The inverse scenario is equally problematic: When visualizations align with real-world values, high accuracy may simply reflect factual recall rather than visual comprehension. Under current evaluation methods, these indistinguishable response sources make it impossible to assess what is genuinely being measured as visualization literacy.

To address these evaluation gaps and more precisely understand LVLMs' visualization interpretation capabilities, we pose the following research questions:

\begin{itemize}
\item \textbf{RQ1 (Performance Assessment)}: How do state-of-the-art proprietary and open-source LVLMs perform on visualization literacy tasks?
\item \textbf{RQ2 (Conflict Resolution)}: When visual information conflicts with factual knowledge, how do LVLMs prioritize between the two sources?
\item \textbf{RQ3 (Preference Steering)}: Can prompt engineering reliably shift LVLMs' prioritization between visual evidence and factual priors under conflict conditions?
\end{itemize}

To investigate these questions, we propose a disentanglement framework that defines two key dimensions, visual correctness and factual correctness, in LVLM-based visualization interpretation (Sec.~\ref{sec:problem_statement}).
These two dimensions have typically been overlooked or treated as a single outcome in prior studies evaluating the performance of LVLMs.
By making these sources of correctness explicit, our framework reveals how prevailing accuracy-based evaluation methods obscure whether reported performance reflects visual interpretation, factual recall, or a mixture of both.

Guided by this framework, we conduct three complementary experiments with \replaced{15}{12} state-of-the-art LVLMs.
First, we establish baseline performance through VLAT \cite{lee2016vlat} and its randomized variant, reVLAT \cite{hong2025llms}, which controls for factual priors while preserving visual encodings.
Second, we introduce the Counterfactual Visualization Literacy Assessment Test (CVLAT), \replaced{a diagnostic benchmark that}{which} systematically constructs visual–factual conflicts to \replaced{measure}{examine} how models arbitrate between visual evidence and stored knowledge \added{when the two diverge}. \added{We position visual--factual arbitration not as an alternative to visualization literacy, but as one of its constituent facets---one that conventional accuracy-based tests entangle with perceptual decoding. By isolating and diagnosing this specific facet, CVLAT complements rather than replaces established assessments such as VLAT and reVLAT, which remain necessary for measuring perceptual decoding itself.}
Finally, we test whether prompt interventions can redirect this prioritization by comparing factual-priority and visual-priority prompts, assessing whether prioritization patterns are intrinsic architectural tendencies or externally steerable preferences.

In summary, our main contributions are:
\begin{itemize}
\item A disentanglement framework that separates visual correctness from factual correctness in LVLM visualization literacy.
\item A comprehensive empirical assessment of \replaced{15}{12} state-of-the-art LVLMs, spanning both proprietary and open-source models.
\item CVLAT: \replaced{A novel diagnostic benchmark for measuring visual--factual arbitration within LVLM visualization literacy.}{A novel literacy assessment test for systematically examining conflicts between visual information and factual knowledge.}
\item An analysis of prompt engineering effectiveness in directing LVLMs' prioritization between visual evidence and factual knowledge.
\end{itemize}
\section{Related Work}
\subsection{Visualization Literacy and Assessment}

Visualization literacy has been defined in various ways in the literature \cite{lee2016vlat, boy2014principled, borner2016investigating}. The most well-known and concise definition of visualization literacy is \textit{the ability and skill to read and interpret visually represented data in and to extract information from data visualizations} \cite{lee2016vlat}. The concept of visualization literacy is connected with the utilization and adoption of visualization and visual analytics tools \cite{maltese2015data, galesic2011graph, bach2021special}. Consequently, there has been a growing interest in quantitatively assessing users' visualization literacy \cite{lee2016vlat, pandey2023mini, boy2014principled, ge2023calvi}. For example, Boy et al. \cite{boy2014principled} proposed an evaluation method based on item response theory (IRT) \cite{baker2001basics} for evaluating individuals' visualization literacy in visualization types including \textit{line graphs}, \textit{bar charts}, and \textit{scatterplots}. Börner et al. \cite{borner2016investigating} assessed the visualization literacy of 273 science museum visitors through familiarity-based questions about various data visualizations. The most widely-adopted assessment tool is the VLAT \cite{lee2016vlat}, which consists of 53 multiple-choice questions with 8 different types of tasks across 12 visualization types. Building upon this research, Pandey and Ottley proposed Mini-VLAT \cite{pandey2023mini}, a concise version of VLAT that reduces the number of questions to 12 while maintaining assessment validity. Ge et al. developed a precise definition of misleaders—decisions made in the construction of visualizations that can lead to conclusions not supported by the data—and proposed CALVI \cite{ge2023calvi} to assess critical thinking about misleading visualizations.

\replaced{While effective for humans, these tests may not transfer to LVLMs: trained on massive datasets that may include the test items themselves, LVLMs blend visual interpretation with pre-trained knowledge in ways that standard assessments were not designed to disentangle. Our work addresses this gap (Sec.~\ref{sec:problem_statement}).}{These assessment tests have proven effective in measuring human visualization literacy. However, we question whether these tests are equally effective for measuring the visualization literacy of Large Vision-Language Models (LVLMs). Unlike humans, LVLMs are trained on massive datasets that may include test materials or similar visualizations, potentially compromising the validity of standard assessment approaches. Additionally, the complex relationship between LVLMs' visual interpretation capabilities and their pre-trained knowledge presents unique evaluation challenges that traditional assessment methods were not designed to address. Our research explores these challenges and introduces a novel approach to LVLM visualization literacy assessment (Sec.~\ref{sec:problem_statement}).}

\subsection{Visualization Literacy in Large Vision Language Models}
In recent years, with advances in the visual understanding of AI models \cite{liu2023visual, park2025s&ui}, the concept of visualization literacy has been extended to LVLMs. As these models continue to evolve, assessing LVLMs' visualization literacy has emerged as a significant new research direction \cite{bendeck2024empirical, hong2025llms, pandey2025benchmarking, li2024visualization, lo2024good}. Across this body of research, the VLAT has emerged as the predominant methodology, providing researchers with a standardized approach to measure and compare visualization literacy across different LVLM architectures. For instance, Bendeck and Stasko \cite{bendeck2024empirical} evaluated GPT-4V's visualization literacy through a series of tests, including the VLAT. Their findings revealed that GPT-4V demonstrated strong capabilities in \textit{identifying trends} and \textit{extreme values}, while showing notable limitations in accurately retrieving specific values from visualizations. Similarly, Li et al. \cite{li2024visualization} assessed visualization literacy in GPT-4o, Claude 3 Opus, and Gemini Pro 1.5 with VLAT \cite{lee2016vlat} and Mini-VLAT \cite{pandey2023mini}. Their study found that these models outperformed the human baseline in \textit{identifying correlations}, \textit{clusters}, and \textit{hierarchical structures}. In more recent research, Pandey and Ottley \cite{pandey2025benchmarking} evaluated the visualization literacy of GPT-4o, Claude 3.5 Sonnet, Gemini 1.5 Pro, and Llama3.2-vision using both VLAT \cite{lee2016vlat} and CALVI \cite{ge2023calvi}. They reported that these LVLMs approached or exceeded human-level performance in tasks such as \textit{trend identification} and \textit{hierarchical structure detection}, but showed poor reliability in interpreting deceptive visualizations.

Notably, Hong et al. \cite{hong2025llms} assessed the visualization literacy of GPT-4V and Gemini using reVLAT—a modified version with randomized data while maintaining the same chart types and task types. Their study found that GPT-4V performed relatively well in \textit{Finding Correlation Trends} and \textit{Making Comparisons} when using \textit{scatterplots}, achieving performance comparable to humans. However, unlike prior research \cite{bendeck2024empirical}, Hong et al. reported that GPT-4V showed weak performance in tasks such as \textit{Finding Extremum}. Importantly, they observed that LVLMs exhibited a strong tendency to rely on their pre-existing knowledge rather than the visualization content when answering questions. These conflicting findings highlight the need for a more nuanced framework to assess whether LVLMs truly understand visualizations or merely leverage their pre-trained knowledge—a gap our research aims to address through empirical investigation (Sec.~\ref{sec:exp2}).

\added{Our work positions itself along two dimensions of prior research. Relative to benchmarking studies such as Hong et al.~\cite{hong2025llms}, we systematically construct visual--factual conflicts grounded in shared factual priors (instead of randomizing data), introduce per-model arbitration metrics with capability controls (instead of reporting a population-level accuracy gap), and test whether prompt-based intervention can steer this reliance. Relative to a complementary line of work that pursues \emph{structural} improvements to LVLM chart understanding via self-training~\cite{huang2025evochart}, reasoning chains~\cite{das2025charts, wu2024chartinsights}, mixture-of-experts architectures~\cite{xu2024chartmoe}, and chart-focused instruction tuning~\cite{masry2025chartgemma}, our work shifts the focus toward diagnosing base-model arbitration, providing a benchmark against which such structurally improved models can be evaluated. We further connect to parallel work that directly compares human and VLM literacy~\cite{verma2025chart} by collecting an $N=30$ Prolific human baseline on CVLAT (Sec.~\ref{sec:cvlat_human}), enabling a direct human--LVLM comparison on the same arbitration items.}

\subsection{Cognitive Bias in Visualization Interpretation}
The influence of human cognitive biases on visualization tasks has been actively researched \cite{wall2017warning, dimara2018task}. Human cognitive biases are psychological mechanisms that systematically distort information during decision-making processes \cite{kahneman2011thinking, tversky1974judgment}. For example, Xiong et al. \cite{xiong2019curse} demonstrated the ``curse of knowledge" bias in data visualization, showing that people with prior knowledge of specific data patterns incorrectly assume others will find the same patterns visually salient, hindering effective communication of insights. These studies underscore that prior knowledge and biases can overshadow raw visual information in humans. 

\replaced{We draw on this body of literature as motivational context for examining whether LVLMs' pretrained knowledge analogously interacts with visual interpretation, treating the human--LVLM analogy as an empirical question rather than a stipulated equivalence. While related phenomena in the context of LLMs and LVLMs are often discussed under various terms such as hallucination, model bias, or reasoning errors, their specific impact on visualization literacy assessment remains largely unexplored.}{Similarly, we posit that analogous biases can manifest in LVLMs' visualization interpretation tasks. The tendency of LVLMs to rely on their pre-existing knowledge rather than actual visualization content, as observed by Hong et al. \cite{hong2025llms}, exemplifies such biases. In the context of LLMs and LVLMs, these phenomena are often referred to using various terms such as hallucination, model bias, or reasoning errors. However, their specific impact on visualization literacy assessment remains largely unexplored. This gap motivates our investigation into the impact of biases in LVLMs' visualization interpretation tasks.}

Recent work on hallucination in LLMs \cite{bang2025hallulens} has proposed taxonomies to classify hallucinations, but these frameworks focus primarily on text-based responses and do not address the unique challenges of visual interpretation tasks. Similarly, Guan et al. \cite{guan2024hallusionbench} introduced a diagnostic benchmark distinguishing between ``language hallucination" and ``visual illusion" in LVLMs, examining how models respond to manipulated images that contradict factual knowledge. However, their work focuses on general image understanding rather than the specific domain of visualization literacy assessment.

Building on these foundations, our work specifically investigates how LVLMs interpret visualizations, examining whether models rely more on visual information or factual knowledge when these two sources align or conflict. By establishing visual correctness and factual correctness as orthogonal dimensions, \replaced{our work makes the visual--factual arbitration explicit and measurable. We use the cognitive-bias literature solely as motivational context for studying LVLM visualization literacy, without assuming shared cognitive mechanisms between humans and LVLMs.}{we apply concepts from cognitive bias research to understand LVLMs' visualization literacy.}
\section{Problem Statement}
\label{sec:problem_statement}

Prior studies that empirically evaluate the visualization literacy of LVLMs~\cite{bendeck2024empirical, hong2025llms, pandey2025benchmarking, li2024visualization, lo2024good} predominantly utilize visualization literacy tests designed for humans~\cite{lee2016vlat, pandey2023mini}.
While these tests enable straightforward accuracy-based analysis for specific visualization and task types, they yield conflicting findings about LVLMs' visualization capabilities.
For example, Bendeck et al.~\cite{bendeck2024empirical} concluded through their replication study that the probability of GPT-4V solving problems by relying on prior knowledge rather than visualization interpretation was low.
In contrast, Hong et al.~\cite{hong2025llms} demonstrated that LVLMs heavily relied on their pre-existing knowledge to answer questions instead of utilizing information from the visualizations. 

Given these conflicting findings, we propose that a thorough evaluation of LVLMs' visualization literacy should extend beyond simple accuracy metrics.
Relying solely on these metrics may mask whether correct answers result from true visual understanding or from leveraging pre-trained knowledge.
Additionally, it may not clarify whether incorrect answers indicate actual failures in interpretation or merely reflect a preference for factual knowledge over visual information.

\begin{figure}[t!]
    \centering
    \includegraphics[width=\columnwidth]{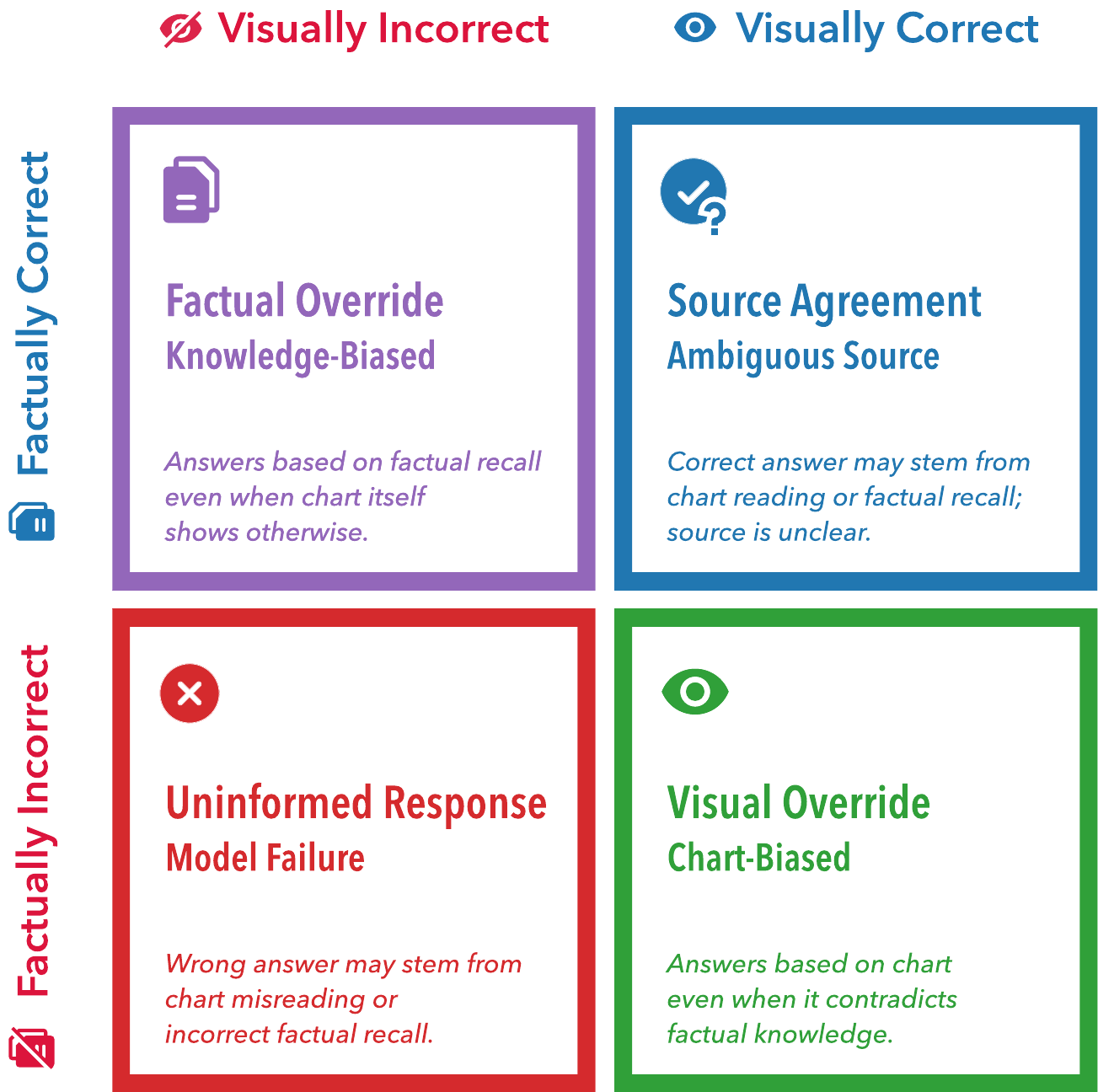}
    \caption{A quadrant framework for evaluating how LVLMs balance visual information and pre-trained knowledge. The framework uses two dimensions: Visual Correctness measuring adherence to visual information, and Factual Correctness measuring alignment with factual knowledge. The four quadrants represent distinct cases of LVLM behavior when interpreting visualizations.}
    \label{fig:quadrant_framework}
\end{figure}

We therefore propose a two-dimensional framework that disentangles visual interpretation from factual knowledge (Figure~\ref{fig:quadrant_framework}).
Rather than evaluating responses along a single correct/incorrect axis, we assess them along two independent dimensions: visual correctness (whether responses align with the presented visualization) and factual correctness (whether responses align with real-world facts). Throughout our discussion, we use the terms ``factual knowledge" and ``prior knowledge" interchangeably to refer to the information that LVLMs have acquired during pre-training, which exists independently of the visual information.

\begin{figure}[tb]
\centering
\includegraphics[width=\columnwidth]{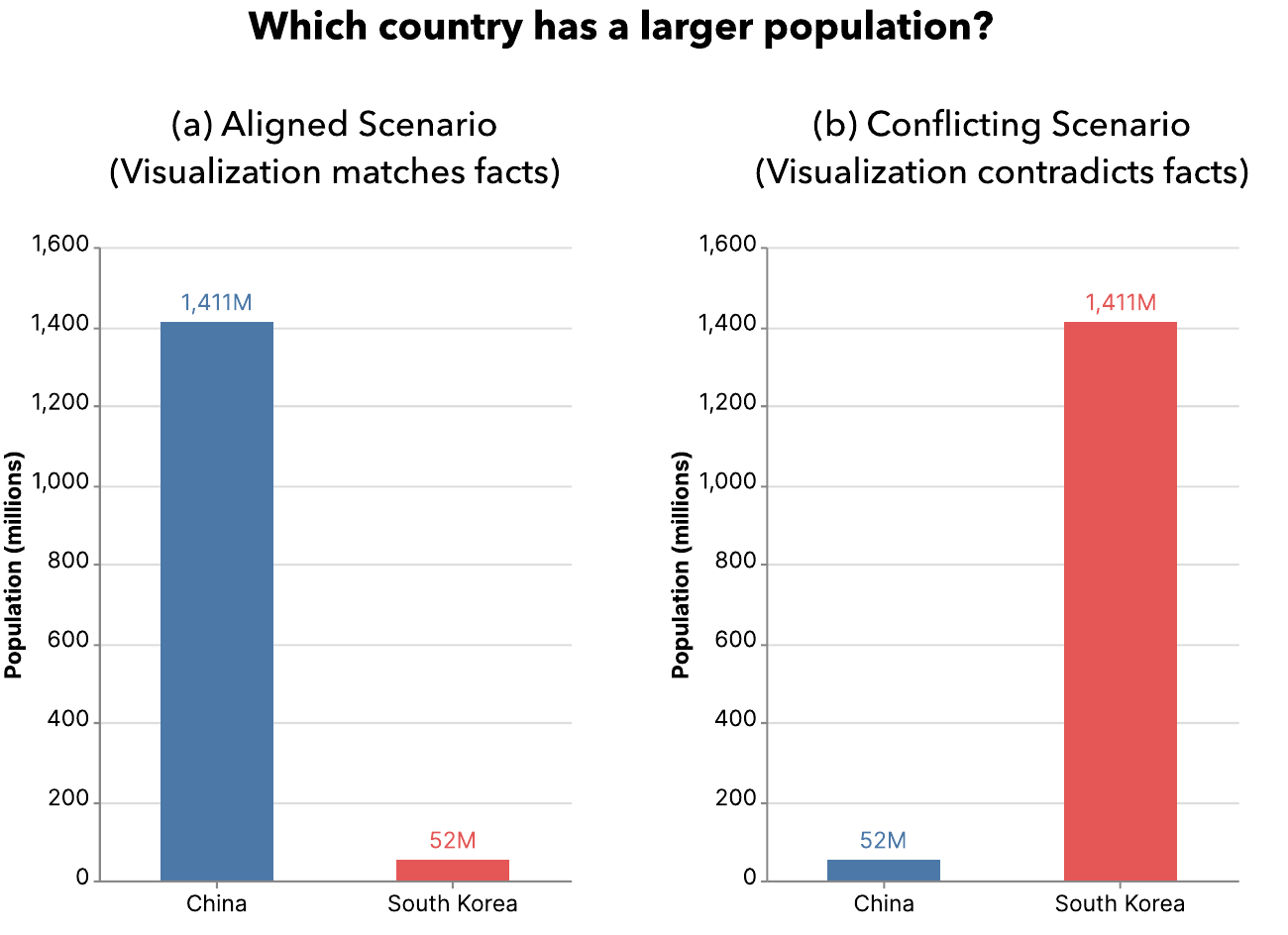}
\caption{Population comparison bar charts demonstrating aligned versus conflicting scenarios when asked ``Which country has a larger population?". (a) Aligned scenario where visualization matches factual knowledge (China $>$ South Korea): answering ``China" falls into the Source Ambiguity (\textbf{VC}$\wedge$\textbf{FC}) quadrant, as we cannot determine if the model read the chart or recalled facts. (b) Conflicting scenario presenting counterfactual data where the visualization shows South Korea $>$ China: answering ``China" falls into the Factual Override (\textbf{VI}$\wedge$\textbf{FC}) quadrant (prioritizing factual knowledge over visual information), while answering ``South Korea" falls into the Visual Override (\textbf{VC}$\wedge$\textbf{FI}) quadrant (correctly interpreting the visualization despite contradicting facts).}
\label{fig:two_scenarios}
\end{figure}

This framework reveals critical patterns in LVLM behavior that single-axis accuracy metrics cannot capture.
By analyzing responses across both dimensions, we identify two primary cases:

\textbf{Aligned cases} (Figure~\ref{fig:two_scenarios}a) occur when visualization and factual knowledge agree:
\begin{itemize}
\item \textbf{Source Ambiguity} --- Visually Correct (\textbf{VC}) and Factually Correct (\textbf{FC}): High accuracy may stem from either visual interpretation or knowledge recall, potentially inflating visualization literacy scores and making it unclear how much of the performance reflects actual visualization literacy.
\item \textbf{Model Failure} --- Visually Incorrect (\textbf{VI}) and Factually Incorrect (\textbf{FI}): 
This clearly indicates failure, but we cannot determine whether the failure stems from misinterpreting the visualization, incorrect knowledge, or both.
\end{itemize}

\textbf{Conflicting cases} (Figure~\ref{fig:two_scenarios}b) emerge when visualization contradicts facts:
\begin{itemize}
\item \textbf{Visual Override} --- Visually Correct (\textbf{VC}) and Factually Incorrect (\textbf{FI}): Following counterfactual visualizations suggests that the model relied on visual information rather than factual knowledge, prioritizing visual information over factuality.
\item \textbf{Factual Override} --- Visually Incorrect (\textbf{VI}) and Factually Correct (\textbf{FC}): Choosing facts over visualization may indicate either interpretation failure or knowledge prioritization, potentially underestimating visualization literacy.
\end{itemize}

This framework clarifies why existing evaluations generate inconsistent findings: they cannot distinguish between these fundamentally different response mechanisms in our two-dimensional space. We therefore designed three complementary experiments: Section~\ref{sec:exp1} establishes baseline performance using VLAT and reVLAT; Section~\ref{sec:exp2} introduces CVLAT to explicitly test conflicting visual-factual cases; Section~\ref{sec:exp3} explores whether prompt engineering can shift models' prioritization between visual evidence and factual knowledge.

\section{Experiment One: Evaluating LVLMs' visualization literacy}
\label{sec:exp1}

In our first experiment, we address \textbf{RQ1}: What is the visualization literacy performance of state-of-the-art LVLMs, including both proprietary and open-source models?

Following prior methodologies \cite{bendeck2024empirical, hong2025llms, pandey2025benchmarking}, we assess \replaced{a state-of-the-art LVLM suite}{12 LVLMs} using VLAT \cite{lee2016vlat} and reVLAT \cite{hong2025llms}.
\replaced{Our evaluation covers the latest proprietary releases and recent open-source families (Table~\ref{tab:models}), many of which have not yet been systematically evaluated for visualization literacy.}{We include recent models (e.g., GPT-4.1, Claude-3.7, and Llama4) that have not been systematically evaluated for visualization literacy, extending the benchmarks from prior studies.}

\begin{table}[ht]
\centering
\caption{Selected LVLMs for Visualization Literacy Assessment}
\label{tab:models}
\begin{threeparttable}
\begin{tabular}{llc}
\toprule
\textbf{Model} & \textbf{Developer} & \textbf{Parameter Size} \\
\midrule
\multicolumn{3}{l}{\textit{Proprietary Models}} \\
\midrule
GPT-5.5 & OpenAI & -$^\dagger$ \\
Claude-Opus-4.7 & Anthropic & -$^\dagger$ \\
Claude-Sonnet-4.6 & Anthropic & -$^\dagger$ \\
Claude-Haiku-4.5 & Anthropic & -$^\dagger$ \\
Gemini-3.1-Pro & Google & -$^\dagger$ \\
Gemini-3.1-Flash-Lite & Google & -$^\dagger$ \\
Grok-4.3 & xAI & -$^\dagger$ \\
Grok-4.20 & xAI & -$^\dagger$ \\
\midrule
\multicolumn{3}{l}{\textit{Open-source Models}} \\
\midrule
Llama4-Maverick & Meta & 400B (17B active)\textsuperscript{*} \\
Llama4-Scout & Meta & 109B (17B active)\textsuperscript{*} \\
Gemma-4-31B & Google & 31B \\
Gemma-4-26B-A4B & Google & 26B (4B active)\textsuperscript{*} \\
Qwen3-VL-235B & Alibaba & 235B (22B active)\textsuperscript{*} \\
Qwen3-VL-32B & Alibaba & 32B \\
Qwen3-VL-8B & Alibaba & 8B \\
\bottomrule
\end{tabular}
\begin{tablenotes}
\fontsize{7pt}{7pt}\selectfont
\setlength{\leftskip}{-1em}
\item $^\dagger$ Parameter sizes for proprietary models are not disclosed due to their proprietary architectures.
\item \replaced{\textsuperscript{*} Active parameters for Mixture-of-Experts (MoE) models.}{* Active parameters. Total parameters for Mixture of Experts (MoE) models: 400B for Llama4-Maverick and 109B for Llama4-Scout.}
\end{tablenotes}
\end{threeparttable}
\end{table}

\subsection{Model Selection}

Our study aims to balance reproducibility with comprehensive coverage of current models.
For proprietary models, we establish minimum reproducibility criteria by selecting only those that support temperature adjustment, which allows consistent result generation across runs.
\deleted{Consequently, we exclude models like OpenAI's o1, o2, and o3 from our candidate pool due to their lack of temperature control functionality.} \added{For models that additionally expose a reasoning-effort or extended-thinking control, we fix this control to its lowest available setting. Because extended internal reasoning is a known source of run-to-run variability even at temperature = 0~\cite{atil2024non}, the lowest-effort setting provides the more reproducible configuration.}

Table~\ref{tab:models} presents the \replaced{15 LVLMs}{12 LVLMs} selected for our evaluation. \added{The exact API snapshots and local Hugging Face checkpoints used are documented in Appendix~D.}
We choose to evaluate base models rather than fine-tuned variants for several reasons.
First, using base models provides a clearer picture of the foundational visual literacy capabilities inherent to each architecture without task-specific optimizations that might obscure underlying limitations.
Second, this approach ensures fair comparison across model families, as fine-tuning techniques and datasets vary widely between research teams and commercial providers, potentially introducing confounding factors into our analysis.
Third, base models represent the most accessible versions for broader research communities, making our findings more generalizable and applicable across various downstream applications where custom fine-tuning may not be feasible.
Finally, we deliberately include a range of open-source models such as \replaced{Llama4, Gemma-4, and Qwen3-VL}{Gemma3, Llama, and Qwen2.5-VL} to provide a more holistic perspective on visualization literacy capabilities across the LVLM landscape.

\subsection{Prompt Design}

We design two prompt conditions for our evaluation (see Appendix~A for full prompt texts). The \textbf{Normal prompt}, based on Hong et al.'s \cite{hong2025llms} approach, instructs models to provide direct answers and select `Omit' when uncertain, without requiring reasoning. We intentionally keep this prompt minimal to avoid confounding the measurement of baseline visualization literacy capabilities.

The \textbf{Explain prompt} incorporates chain-of-thought prompting \cite{wei2022chain} informed by semantic content frameworks from visualization research \cite{lundgard2021accessible, ko2024natural}. It guides models through three stages aligned with Lundgard and Satyanarayan's framework \cite{lundgard2021accessible}: (1) describing visual attention patterns, (2) extracting data values corresponding to elemental encoding recognition, and (3) explaining calculations and interpretations encompassing statistical relationships and perceptual patterns.

\subsection{Experimental Design and Validity}

\subsubsection{Test Set Selection and Rationale}
Using the quadrant framework introduced in Section~\ref{sec:problem_statement}, we first examine VLAT, which constructs visualizations using data that generally aligns with real-world knowledge.
Under this setting, incorrect responses fall into the Model Failure (\textbf{VI}$\wedge$\textbf{FI}) case: The model clearly fails, but we cannot determine whether the error stems from visual misinterpretation or incorrect factual recall.
Similarly, correct responses fall into the Source Ambiguity (\textbf{VC}$\wedge$\textbf{FC}) case, where we cannot distinguish genuine visual interpretation from pretrained knowledge recall, potentially inflating visualization literacy scores.

To complement this assessment, we adapt reVLAT as our second test.
reVLAT preserves VLAT's chart and task types but replaces underlying data with randomized values\replaced{, which break the alignment between visualizations and the original real-world data}{, ensuring visualizations either contradict or bear no relation to real-world facts}.
Under this setting, correct responses \added{typically} fall into the Visual Override (\textbf{VC}$\wedge$\textbf{FI}) case, indicating genuine visualization interpretation. However, incorrect responses remain ambiguous—they may indicate true failures to interpret the visualization or correspond to the Factual Override (\textbf{VI}$\wedge$\textbf{FC}) case, where factual priors outweigh visual evidence. Consequently, reVLAT may underestimate visualization literacy by not distinguishing between these two error sources.

\subsubsection{Experimental Protocol and Evaluation Methodology}
VLAT and reVLAT each contain 53 multiple-choice questions, with varying numbers of answer options (3, 4, or 5), including an `Omit' option.
To control for potential ordering effects reported in prior research \cite{hong2025llms, pandey2025benchmarking}, we generate 120 answer-ordering variants per question. For 5-option questions, this includes all unique permutations. For questions with fewer options (3 or 4), we proportionally repeat each unique permutation to reach 120 variants.
Our experimental design yields 6,360 question variations per model for each prompt-test combination (120 permutations $\times$ 53 questions).
With 2 test types and 2 prompt conditions, each LVLM processes 25,440 distinct instances.
Across the \replaced{15}{12} LVLMs, this produces \replaced{381{,}600}{305{,}280} total trials, providing robust statistical power for our comparative analysis.

Given the scale of our experiment, we adopt an efficient and consistent evaluation pipeline.
Following established practices \cite{guan2024hallusionbench, liu2023g}, we use \replaced{GPT-5.4-nano}{GPT-4o} to extract letter-based answers from model responses.
The extraction process employs structured prompts to identify the selected option (i.e., letter-based answer) from each model's response text, accommodating varied response formats in which models provide explanations, qualifiers, or expressions of uncertainty alongside their answers.
To ensure consistency and robustness in answer extraction, we conduct five independent extraction runs with \replaced{GPT-5.4-nano}{GPT-4o} (temperature=0) and apply majority voting to determine the final answer for each trial. \added{The same extraction pipeline is reused in all subsequent experiments and capability-reference conditions; across all $813{,}600$ resulting trials in the full study suite, the pipeline produced a structured letter selection for every response, with no unparseable extractions.}

\subsubsection{Analysis Methodology}
For \replaced{each model and experimental condition we report mean accuracy and standard deviation across questions, where per-question accuracy is computed as the mean over the 120 permutations to control for}{statistical analysis, we compute mean accuracy scores and standard deviations for each model and experimental condition.
To control for ordering effects, accuracy is first averaged at the question level across all permutations. We then apply bootstrap resampling (10,000 iterations) to these question-level means to estimate 95\% confidence intervals.
This two-stage aggregation and resampling procedure accounts for both question-level variability and} ordering effects. \added{We summarize Experiment 1 dispersion with standard deviations across the 53 items, whereas Experiments 2 and 3, which test directional hypotheses about arbitration shifts, additionally report bootstrap confidence intervals.}
We also analyze performance across visualization types (e.g., bar charts, line graphs) and task categories (e.g., trend identification, value retrieval) to identify model-specific strengths and limitations. Detailed breakdowns are provided in Appendix~B.

\begin{table*}[t]
\centering
\caption{Performance comparison of LVLMs on visualization literacy tests. Best performances among \replaced{proprietary models and best among open-source models are highlighted in \textbf{\underline{bold and underlined}}.}{proprietary models are highlighted in \textcolor{red}{red}, and best among open-source models are highlighted in \textcolor{blue}{blue}.} The $\Delta$ columns show the performance drop from VLAT to reVLAT.}
\label{tab:results_experiment1}
\begin{threeparttable}
\begin{tabular}{lc|cc|cc|cc}
\toprule
\multirow{2}{*}{\textbf{Model}} & \multirow{2}{*}{\textbf{Provider}} & \multicolumn{2}{c|}{\textbf{VLAT}} & \multicolumn{2}{c|}{\textbf{reVLAT}} & \multicolumn{2}{c}{\textbf{$\Delta$ (VLAT$-$reVLAT)}} \\
\cmidrule{3-8}
& & \textbf{Normal} & \textbf{Explain} & \textbf{Normal} & \textbf{Explain} & \textbf{Normal} & \textbf{Explain} \\
\midrule
\textit{Human Baseline} \cite{lee2016vlat} & -- & \multicolumn{2}{c|}{65.50 (13.30)} & \multicolumn{2}{c|}{--} & \multicolumn{2}{c}{--} \\
\midrule
GPT-5.5 & OpenAI & 80.09 (26.93) & 93.85 (20.69) & 66.19 (34.81) & 87.97 (28.52) & 13.90 & 5.88 \\
Claude-Opus-4.7 & Anthropic & 94.87 (15.38) & 94.70 (17.45) & 88.52 (26.29) & 90.63 (25.59) & 6.35 & 4.07 \\
Claude-Sonnet-4.6 & Anthropic & 89.43 (26.43) & 83.40 (24.71) & 79.62 (36.88) & 73.24 (37.17) & 9.81 & 10.16 \\
Claude-Haiku-4.5 & Anthropic & 77.09 (32.30) & 72.08 (29.57) & 67.61 (38.57) & 64.65 (35.40) & 9.48 & 7.43 \\
Gemini-3.1-Pro & Google & \textbf{\underline{99.73}} (1.73) & \textbf{\underline{98.79}} (4.77) & \textbf{\underline{93.29}} (23.79) & \textbf{\underline{93.21}} (24.19) & 6.44 & 5.58 \\
Gemini-3.1-Flash-Lite & Google & 77.33 (32.52) & 90.19 (23.55) & 69.70 (36.92) & 88.00 (24.63) & 7.63 & 2.19 \\
Grok-4.3 & xAI & 44.36 (28.00) & 77.72 (27.26) & 33.52 (29.92) & 72.09 (29.50) & 10.84 & 5.63 \\
Grok-4.20 & xAI & 37.89 (29.27) & 74.95 (26.46) & 37.66 (32.37) & 70.77 (29.73) & 0.23 & 4.18 \\
\cmidrule{1-8}
\textit{Proprietary Average} & -- & 75.10 & 85.71 & 67.01 & 80.07 & 8.09$^\dagger$ & 5.64$^\dagger$ \\
\midrule
Llama4-Maverick & Meta (Local) & 60.58 (34.77) & 67.75 (37.48) & 49.58 (35.44) & 69.07 (37.43) & 11.00 & -1.32 \\
Llama4-Scout & Meta (Local) & 48.51 (34.87) & 61.37 (36.51) & 40.25 (32.88) & 55.72 (36.81) & 8.26 & 5.65 \\
Gemma-4-31B & Google (Local) & \textbf{\underline{67.77}} (35.80) & 82.23 (33.84) & 63.60 (39.75) & \textbf{\underline{83.95}} (32.70) & 4.17 & -1.72 \\
Gemma-4-26B-A4B & Google (Local) & 58.38 (39.83) & \textbf{\underline{86.24}} (28.67) & \textbf{\underline{63.84}} (35.93) & 82.16 (31.67) & -5.46 & 4.08 \\
Qwen3-VL-235B & Alibaba & 60.03 (35.47) & 78.95 (31.54) & 51.37 (38.28) & 78.43 (33.10) & 8.66 & 0.52 \\
Qwen3-VL-32B & Alibaba (Local) & 67.01 (36.15) & 81.97 (29.10) & 54.48 (38.16) & 79.21 (34.43) & 12.53 & 2.76 \\
Qwen3-VL-8B & Alibaba (Local) & 53.22 (31.50) & 67.41 (32.62) & 38.79 (35.74) & 65.14 (36.30) & 14.43 & 2.27 \\
\cmidrule{1-8}
\textit{Open-source Average} & -- & 59.36 & 75.13 & 51.70 & 73.38 & 9.22$^\dagger$ & 2.62$^\dagger$ \\
\bottomrule
\end{tabular}
\begin{tablenotes}
\fontsize{7pt}{7pt}\selectfont
\item * Values shown as: mean accuracy percentage (standard deviation) for VLAT and reVLAT columns; percentage point differences for $\Delta$ columns.
\item * Human Baseline represents average human performance on VLAT reported by Lee et al. \cite{lee2016vlat}.
\item * Negative $\Delta$ values indicate better performance on reVLAT than VLAT.
\item $^\dagger$ Average of absolute values.
\end{tablenotes}
\end{threeparttable}
\end{table*}
\subsection{Experimental Results}
\subsubsection{Overall Performance Analysis}

Table~\ref{tab:results_experiment1} presents the performance results of \replaced{15}{12} state-of-the-art LVLMs on both VLAT and reVLAT under Normal and Explain prompt conditions.

On VLAT, which contains visualizations generally aligned with real-world data, several models demonstrated strong performance.
Under the Normal prompt condition, \replaced{six proprietary LVLMs exceeded the human baseline of 65.50\%: Gemini-3.1-Pro achieved the highest accuracy at 99.73\%, followed by Claude-Opus-4.7 (94.87\%), Claude-Sonnet-4.6 (89.43\%), GPT-5.5 (80.09\%), Gemini-3.1-Flash-Lite (77.33\%), and Claude-Haiku-4.5 (77.09\%). Among open-source models, Gemma-4-31B (67.77\%) and Qwen3-VL-32B (67.01\%) also approached or exceeded the human baseline.}{three proprietary LVLMs exceeded the human baseline of 65.50\%: Gemini-2.5-Pro-Preview achieved the highest accuracy at 84.01\%, followed by GPT-4.1 (67.34\%) and Gemini-2.5-Flash-Preview (69.25\%). Among open-source models, Llama4-Maverick (60.58\%) and Qwen2.5-VL-72b (60.42\%) achieved the highest performance, approaching the human baseline of 65.50\%.}

\replaced{The Explain prompt condition produced highly heterogeneous effects across models. The two Grok variants showed the most dramatic improvements (Grok-4.20: from 37.89\% to 74.95\%; Grok-4.3: from 44.36\% to 77.72\%), and GPT-5.5 also improved substantially from 80.09\% to 93.85\%. In contrast, the two non-Opus Claude variants exhibited \emph{negative} Explain-vs.-Normal effects on VLAT (Claude-Sonnet-4.6: from 89.43\% to 83.40\%; Claude-Haiku-4.5: from 77.09\% to 72.08\%), indicating that the structured Explain prompt did not benefit these models' accuracy. Meanwhile, Gemini-3.1-Pro and Claude-Opus-4.7, both already near-ceiling under the Normal condition, showed essentially no Explain effect.}{Notably, the Explain prompt condition generally improved performance across most models, with Claude-3.7-Sonnet showing the most dramatic improvement from 58.79\% to 86.95\%, and Gemini-2.5-Pro-Preview reaching 90.05\%. This exceptional improvement in Claude-3.7-Sonnet may be attributed to its hybrid architecture that combines reasoning and non-reasoning capabilities, which may interact favorably with the structured Explain prompt.}

However, performance changed dramatically on reVLAT, which uses randomized data.
Most models experienced substantial performance drops, with GPT-\replaced{5.5 declining from 80.09\% to 66.19\% and Claude-Sonnet-4.6 dropping from 89.43\% to 79.62\%. Among the top-performing proprietary models, Gemini-3.1-Pro and Claude-Opus-4.7 retain remarkably high reVLAT accuracy (93.29\% and 88.52\%, drops of only 6.44pp and 6.35pp, respectively). Grok-4.20 shows the smallest absolute drop in the suite (0.23pp from 37.89\% to 37.66\%), though from a low baseline. Among open-source models, Gemma-4-26B-A4B (63.84\%) and Gemma-4-31B (63.60\%) achieve the highest reVLAT Normal accuracies}{4o declining from 59.92\% to 36.48\% and Qwen2.5-VL-72b dropping from 60.42\% to 44.97\%. Interestingly, despite being a lightweight model, Qwen2.5-VL-7b achieved the highest accuracy among open-source models on reVLAT under Normal prompt condition (50.88\%)}.
This substantial performance gap between VLAT and reVLAT reflects the ambiguity in evaluation described in our quadrant framework (Sec.~\ref{sec:problem_statement}). On VLAT, where visual and factual information align, high accuracy may stem from either genuine visual interpretation or factual recall (Source Ambiguity (\textbf{VC}$\wedge$\textbf{FC})); on reVLAT, where factual priors no longer apply, the performance drop suggests that some models relied on factual knowledge rather than visual interpretation when answering VLAT questions.

When examining group-level performance, proprietary models achieved higher average accuracies than open-source models: \replaced{75.10\% (Normal) and 85.71\% (Explain) on VLAT, compared to 59.36\% and 75.13\% for open-source models. However, individual comparisons reveal a more nuanced picture: several open-source models (Gemma-4-31B, Qwen3-VL-32B, Llama4-Maverick) match or exceed the lower-performing proprietary models like Grok-4.20 and Grok-4.3.}{65.26\% (Normal) and 76.76\% (Explain) on VLAT compared to 50.76\% and 56.34\% for open-source models. However, this difference is largely driven by Gemini-2.5-Pro-Preview's exceptional performance. Individual comparisons reveal a more nuanced picture---several open-source models (Llama4-Maverick, Qwen2.5-VL-72b, and Qwen2.5-VL-7b) matched or exceeded the performance of lower-performing proprietary models like Grok-2-Vision and GPT-4o.}

\replaced{The pattern of performance drops from VLAT to reVLAT depends on the prompt condition. Under the Normal prompt, the proprietary group exhibits a slightly smaller drop (8.09pp) than the open-source group (9.22pp). Under the Explain prompt, however, the open-source group exhibits a markedly smaller drop (2.62pp) than the proprietary group (5.64pp). Notably, several open-source models actually show an inverse pattern, performing better on reVLAT than VLAT (e.g., Gemma-4-31B: 82.23\% on VLAT, 83.95\% on reVLAT; and Llama4-Maverick: 67.75\% on VLAT, 69.07\% on reVLAT). Within the proprietary group, Gemini-3.1-Pro and Claude-Opus-4.7 stand out for their robustness to data randomization, showing minimal drops of 6.44pp and 6.35pp under the Normal-prompt condition, respectively. These divergent group-level patterns suggest that some recent models show smaller accuracy drops under data randomization, consistent with reduced dependence on aligned factual priors, though, as Experiment 2 will show, such robustness is conceptually distinct from visual-factual arbitration.}{Interestingly, proprietary models as a group exhibited larger average performance drops between VLAT and reVLAT (13.78pp for Normal, 12.57pp for Explain) compared to open-source models (7.77pp and 7.28pp respectively). While notable exceptions exist---Gemini-2.5-Pro-Preview demonstrated remarkable robustness (5.90pp and 10.80pp) among proprietary models, and Qwen2.5-VL-72b showed exceptionally large drops (15.45pp for Normal, 20.85pp for Explain) despite being open-source---the overall pattern remains consistent. The majority of proprietary models (5 out of 6) showed substantial performance degradation, suggesting that higher-performing models may achieve their superior VLAT scores partly through leveraging pre-trained factual knowledge rather than pure visual interpretation capabilities.}

Prior research \cite{hong2025llms} attributed these performance differences to LVLMs' reliance on pre-existing knowledge rather than visual interpretation. Our quadrant framework provides a more precise account of this phenomenon. The observed performance patterns reveal two key insights:

First, on VLAT, where visual information aligns with real-world facts and thus responses fall into the Source Ambiguity (\textbf{VC}$\wedge$\textbf{FC}) and Model Failure (\textbf{VI}$\wedge$\textbf{FI}) cases, we cannot definitively determine whether high accuracy stems from genuine visualization interpretation or from factual recall. As a result, VLAT scores may overstate visualization literacy by masking the underlying basis of correctness.

Second, the dramatic performance drop on reVLAT raises an interpretive challenge. Although such declines might suggest weak visualization interpretation abilities, they may alternatively indicate that models correctly interpreted the visualizations but prioritized their factual knowledge when generating responses, corresponding to the Factual Override (\textbf{VI}$\wedge$\textbf{FC}) case. This distinction is crucial: if models are indeed interpreting visualizations correctly but defaulting to factual knowledge under conflict, then current accuracy-based evaluation methods may systematically underestimate their true visualization literacy capabilities.

These findings motivate our subsequent experiments, which systematically examine how LVLMs resolve conflicts between visual information and factual knowledge under controlled conditions.

\section{Experiment Two: Assessing LVLMs' Visualization Literacy with Counterfactual Visualizations} 
\label{sec:exp2}

In our second experiment, we address \textbf{RQ2}: How do LVLMs respond when visual information conflicts with factual knowledge, and how do conflicts shape their information prioritization?

\replaced{This connects to McNutt et al.'s `Visualization Mirages' \cite{mcnutt2020surfacing}, where visual encodings interact with prior knowledge to cause misinterpretation. Whereas Mirages characterize human failures, we ask how LVLMs resolve visual--factual conflicts.}{This investigation is related to the notion of `Visualization Mirages' introduced by McNutt et al. \cite{mcnutt2020surfacing}---which describes situations where interaction between visual encodings and prior knowledge can lead to misinterpretation of visualizations.
While Visualization Mirages primarily characterize human interpretive failures, they motivate a complementary line of inquiry for LVLMs: understanding how conflicts between visual input and factual priors influence model responses, and whether such conflicts are resolved through visual interpretation or factual override.}

Previous studies \cite{bendeck2024empirical, pandey2025benchmarking, lo2024good} have evaluated LVLMs' visualization literacy using deceptive visualizations with different types of misleaders, including those arising from data curation \cite{kim2003taxonomy}, data wrangling \cite{cockburn2018hark, lo2022misinformed}, and visualization design \cite{correll2018looks, pandey2015deceptive}.
While these approaches reveal important facets of visualization literacy, they do not isolate the specific challenge we address. Particularly, they did not explicitly examine how LVLMs arbitrate between visual evidence and factual knowledge when the two are in conflict during the visualization interpretation process.

This distinction is essential for accurately assessing visualization literacy, as it gives rise to two problematic assessment cases:
\begin{enumerate}
\item \textbf{Aligned visual-factual cases}: When visualizations align with factual knowledge, LVLMs may answer correctly by relying on prior knowledge rather than interpreting the visualization itself. Such responses fall into the Source Ambiguity (\textbf{VC}$\wedge$\textbf{FC}) case and can artificially inflate reported visualization literacy scores.
\item \textbf{Conflicting visual-factual cases}: When visualizations contradict factual knowledge, LVLMs may generate factually correct answers while ignoring visual information. These responses correspond to the Factual Override (\textbf{VI}$\wedge$\textbf{FC}) case and can lead to underestimation of true visualization literacy.
\end{enumerate}

We focus on these two cases because they are the primary sources of systematic bias in accuracy-based evaluation; the remaining cases are either diagnostically unambiguous or non-informative.

To investigate these cases systematically, we introduce the Counterfactual Visualization Literacy Assessment Test (CVLAT), which uses counterfactual visualizations that deliberately conflict with widely-known facts. By inducing visual-factual conflicts, \replaced{CVLAT lets us estimate aggregate visual--factual arbitration tendencies under controlled conflict}{CVLAT allows us to directly assess whether LVLMs prioritize visual evidence or default to factual knowledge when the two sources diverge}.

\subsection{Experimental Design and Methodology}
\subsubsection{Design of CVLAT and Rationale}

For a systematic assessment of how LVLMs prioritize between visual information and factual knowledge, we design the Counterfactual Visualization Literacy Assessment Test (CVLAT), adapting VLAT with deliberately counterfactual visualizations.

Although visualization literacy tests with randomized data such as reVLAT \cite{hong2025llms} provide useful controls, they are not well suited to our specific research goals.
Data randomization does not guarantee that (1) models have prior knowledge about the topic domain of interest, (2) meaningful conflicts between visual and factual information occur, or (3) such conflicts can be used to analyze prioritization behavior.
For example, reVLAT includes arbitrary height-weight scatterplots for which models lack any factual reference point.
Without verifiable facts to contradict, models have no choice but to follow the visualization, preventing assessment of visual-factual prioritization. \added{Unlike reVLAT, which randomizes chart values and thereby weakens the alignment with factual priors, CVLAT \emph{preserves} shared factual priors and deliberately constructs visual--factual conflicts. This design allows us to operationalize visual-factual arbitration as a primary, controlled variable, rather than an incidental by-product of factual misalignment.}

To address these limitations, CVLAT follows three key design principles:

First, we replace VLAT's original datasets with data drawn from domains strongly grounded in widely shared factual knowledge, including economic indicators, political demographics, and natural phenomena. This ensures that models are likely to possess relevant factual priors that can conflict with visual encodings.

Second, we exclude purely perceptual tasks such as \textit{Find Anomalies} and \textit{Find Clusters}.
These tasks primarily involve pattern recognition rather than data interpretation or factual knowledge integration. This refinement reduces the original 53 VLAT questions to 48 questions that focus on data interpretation, where visual-factual conflicts can meaningfully arise.

Third, we structure answer options to explicitly detect information prioritization:
\begin{itemize}
    \item a visually correct option derived from the counterfactual visualization
    \item a factually correct option based on real-world knowledge
    \item an `Omit' option to allow uncertainty
    \item distractor options for 4- and 5-choice questions
\end{itemize}

\added{The option structure together with the 120-permutation sweep instantiates the forced-choice paradigm of psychophysics~\cite{green1966signal, macmillan2005detection} and the syllogistic belief-bias paradigm of Trippas et al.~\cite{trippas2014using}: contradictory signals are deliberately presented and the model must choose among labeled options including an explicit ``Omit'' escape, so that systematic per-model leans across all trials are interpretable as preferences.}

\subsubsection{\added{Human baseline study (design)}}\label{sec:cvlat_human}
\added{To calibrate CVLAT difficulty against the established VLAT human baseline, we administered the same 48 CVLAT items to $N=30$ Prolific participants using the same multiple-choice format and correction-for-guessing scoring as the original VLAT study~\cite{lee2016vlat}. Items were presented one per page with option order randomized per participant. Participants were not informed of the counterfactual construction. Results of this calibration are reported in Sec.~\ref{sec:exp2_results} (alongside model results) and Appendix~C.}

\subsubsection{Experimental Protocol and Evaluation Methodology}
The Counterfactual Visualization Literacy Assessment Test (CVLAT) comprises 48 multiple-choice questions with varying option counts (3, 4, or 5 options), each including an `Omit' choice.
Following the methodology of Experiment 1, we address potential ordering effects by systematically generating all possible permutations based on the 5-option questions.
This results in each model processing 5,760 distinct question variations (120 permutations $\times$ 48 questions) using the Normal prompt condition from Experiment 1\deleted{, yielding 69,120 total trials across all 12 LVLMs}\added{, yielding 86{,}400 total trials across all 15 LVLMs}.

\subsubsection{\added{Capability references}}\label{sec:cvlat_gt_redundancy}
\added{Alongside CVLAT, we administer two control conditions derived from the same item set: the \emph{anonymized visual baseline} ($V_{\text{anon}}$) and the \emph{Q-only} condition. The anonymized visual baseline presents the same counterfactual chart with domain-identifying cues replaced by neutral placeholders, removing the factual signal so that the resulting response distribution measures pure chart-reading capability. Symmetrically, the Q-only condition asks the same domain question without any accompanying chart, allowing its factual-correct rate ($F_Q$) to estimate the model's factual-prior availability. Because both conditions repurpose the exact CVLAT items rather than introducing an external control set, they disentangle chart-reading proficiency from factual-prior reliance without content-based confounds.}

\begin{table*}[ht]
\centering
\caption{CVLAT evaluation results\replaced{. \textbf{Anon V\%} (visual-selection rate on the anonymized visual baseline) and \textbf{Q-only \%} (text-only factual-knowledge accuracy) are capability references used to normalize VF and FA, respectively. The \textbf{False} column reports the share of responses that are neither visually nor factually correct (i.e., distractor selections or Omit), and Visual, Factual, and False sum to 100\%. VFRI is the relative preference between VF and FA (range $-1$ purely factual to $+1$ purely visual). The 95\% CI column reports the percentile bootstrap interval ($B = 10{,}000$) of the per-question VFRI. Group labels are descriptive, and models whose 95\% CI includes zero should be interpreted as near-boundary cases. VF and FA are capability-normalized ratios and are not bounded above by 1, which is why the low-capability Grok rows exceed 1. VFRI is the bounded $[-1,+1]$ summary index}{ across 12 LVLMs showing response distribution and information prioritization metrics. VF Score measures visual adherence, FA Score measures factual reliance, and VFRI indicates relative preference (range: $-1$ for purely factual to $+1$ for purely visual). Models are grouped by their predominant information prioritization pattern: factual knowledge-oriented (10 models) and visualization-oriented (2 models)}.}
\label{tab:cvlat_results}
\begin{tabular}{l|cc|ccc|cccc}
\toprule
\multirow{2}{*}{\textbf{Model}} & \added{\textbf{Anon V}} & \added{\textbf{Q-only}} & \multicolumn{3}{c|}{\textbf{Response Dist.\ (\%)}} & \multicolumn{4}{c}{\textbf{Scores}} \\
\cmidrule{4-10}
& \added{(\%)} & \added{(\%)} & \textbf{Visual} & \textbf{Factual} & \textbf{False} & \textbf{VF} & \textbf{FA} & \textbf{VFRI} & \added{\textbf{95\% CI}} \\
\midrule
\multicolumn{10}{l}{\textit{Factual knowledge-oriented models \added{(point-estimate VFRI $< 0$)}}} \\
\midrule
Grok-4.20 & 35.83 & 57.64 & 17.45 & 57.76 & 24.79 & 0.543 & 1.068 & -0.594 & [$-0.78$, $-0.37$] \\
Qwen3-VL-32B & 73.06 & 62.81 & 38.12 & 50.68 & 11.20 & 0.422 & 0.859 & -0.496 & [$-0.72$, $-0.25$] \\
Qwen3-VL-235B & 59.60 & 62.90 & 27.57 & 53.85 & 18.58 & 0.352 & 0.707 & -0.358 & [$-0.61$, $-0.10$] \\
Llama4-Scout & 60.69 & 76.65 & 25.30 & 55.05 & 19.65 & 0.380 & 0.904 & -0.339 & [$-0.52$, $-0.15$] \\
Grok-4.3 & 33.70 & 63.66 & 17.19 & 59.76 & 23.06 & 1.070 & 1.437 & -0.293 & [$-0.48$, $-0.10$] \\
Llama4-Maverick & 72.03 & 84.83 & 33.25 & 54.18 & 12.57 & 0.386 & 0.541 & -0.106 & [$-0.37$, $+0.15$] \\
\midrule
\multicolumn{10}{l}{\textit{Visualization-oriented models \added{(point-estimate VFRI $\geq 0$)}}} \\
\midrule
Qwen3-VL-8B & 46.20 & 62.73 & 32.62 & 47.20 & 20.17 & 0.880 & 0.545 & +0.048 & [$-0.27$, $+0.36$] \\
Claude-Haiku-4.5 & 82.97 & 74.65 & 49.18 & 43.40 & 7.41 & 0.552 & 0.671 & +0.120 & [$-0.15$, $+0.39$] \\
Gemma-4-31B & 70.94 & 73.80 & 45.97 & 38.18 & 15.85 & 0.585 & 0.413 & +0.172 & [$-0.11$, $+0.45$] \\
Gemma-4-26B-A4B & 82.47 & 77.52 & 56.28 & 31.37 & 12.34 & 0.631 & 0.368 & +0.208 & [$-0.07$, $+0.47$] \\
GPT-5.5 & 75.26 & 84.70 & 49.01 & 38.33 & 12.66 & 0.644 & 0.394 & +0.247 & [$-0.01$, $+0.50$] \\
Claude-Sonnet-4.6 & 88.11 & 84.84 & 63.54 & 30.82 & 5.64 & 0.864 & 0.344 & +0.310 & [$+0.05$, $+0.56$] \\
Gemini-3.1-Flash-Lite & 75.36 & 80.94 & 50.05 & 36.72 & 13.23 & 0.586 & 0.329 & +0.325 & [$+0.04$, $+0.59$] \\
Claude-Opus-4.7 & 91.70 & 83.21 & 79.44 & 13.65 & 6.91 & 0.792 & 0.235 & +0.669 & [$+0.44$, $+0.87$] \\
Gemini-3.1-Pro & 94.25 & 88.87 & 90.80 & 6.11 & 3.09 & 0.987 & 0.043 & +0.892 & [$+0.77$, $+0.98$] \\
\bottomrule
\end{tabular}
\end{table*}

\subsubsection{Evaluation Metrics}
We define three complementary evaluation metrics to quantify how LVLMs prioritize visual information and factual knowledge under conflict:

\begin{itemize}
    \item \textbf{Visualization Fidelity Score (VF Score)}: Measures adherence to visual information
    \item \textbf{Factual Alignment Score (FA Score)}: Measures reliance on factual knowledge
    \item \textbf{Visual-Factual Reliance Index (VFRI)}: Captures relative preference between visual and factual sources
\end{itemize}

\added{In the definitions below, $V_{\text{CVLAT}}$ and $F_{\text{CVLAT}}$ denote visual-aligned and factual-aligned accuracy in CVLAT, while $V_{\text{anon}}$ and $F_Q$ serve as reference scores for capability normalization.}

\noindent\textbf{Correction for Guessing}\\
To address the issue of guessing in multiple-choice questions\deleted{ that CVLAT employs}, we apply a correction-for-guessing formula adapted from educational assessment literature~\cite{diamond1973correction, frary1988formula, thorndike1991measurement}. For each question $i$, let $S_i$ denote the \replaced{rate of the target response, $W_i$ the rate of distractor responses, $C_i$ the number of answer options, and $D_i$ the number of distractor options.}{proportion of correct responses, $W_i$ the proportion of incorrect responses that are neither visually nor factually correct, and $C_i$ the total number of answer options. Since visual and factual options are the two primary answer types, we correct for guessing among the remaining $C_i - 2$ distractors.} The corrected score is:

\begin{equation}
\label{eq:score}
\text{Score}_i = \max\left(0, S_i - \frac{W_i}{D_i}\right).
\end{equation}

This formula penalizes random guessing by subtracting the expected contribution of \replaced{distractor responses}{false responses distributed across distractor options}, ensuring that chance-level performance yields a score near zero. \added{The number of distractors $D_i$ depends on the specific condition. In CVLAT, both visual-correct and factual-correct options are treated as focal response categories by construction, yielding $D_i = C_i - 2$. In contrast, for the capability-reference conditions ($V_{\text{anon}}$, Q-only), only one focal target category exists, yielding $D_i = C_i - 1$.}

\noindent\textbf{Visualization Fidelity Score (VF Score)}\\
The VF Score measures the extent to which LVLMs follow visual information when it contradicts factual knowledge, corresponding to the Visual Override (\textbf{VC}$\wedge$\textbf{FI}) case in our framework. \replaced{We apply Equation~\ref{eq:score} on CVLAT with $S_i$ set to the visual-correct rate to obtain $V_{\text{CVLAT},i}$, and to the anonymized visual baseline under the same $S_i$ setting to derive the chart-reading capability reference $V_{\text{anon},i}$.}{For each question $i$, we apply Equation~\ref{eq:score} with $S_i$ being the proportion of visually correct responses to obtain the guessing-corrected score $VF_i$.} \added{To prevent capability deficits (e.g., a model that simply cannot read the chart) from being mistaken for factual bias, we normalize $V_{\text{CVLAT},i}$ by the capability reference $V_{\text{anon},i}$:
\begin{equation}\label{eq:vf_norm}
VF_i = \frac{V_{\text{CVLAT},i}}{V_{\text{anon},i} + \varepsilon},
\end{equation}
where $\varepsilon = 10^{-6}$ prevents division by zero when the capability reference is near zero.} The overall VF Score is\deleted{ computed as} the mean of $VF_i$ across all questions\deleted{, yielding a value in the range [0, 1]}.

\noindent\textbf{Factual Alignment Score (FA Score)}\\
The FA Score quantifies the extent to which LVLMs prioritize factual knowledge over contradictory visual information, corresponding to the Factual Override (\textbf{VI}$\wedge$\textbf{FC}) case in our framework. \replaced{We apply Equation~\ref{eq:score} on CVLAT with $S_i$ set to the factual-correct rate to obtain $F_{\text{CVLAT},i}$. Following the same formulation, we evaluate the Q-only condition to obtain the factual-prior availability reference $F_{Q,i}$.}{For each question $i$, we apply Equation~\ref{eq:score} with $S_i$ defined as the proportion of factually correct responses, generating a guessing-corrected score $FA_i$.} \added{Symmetrically to the visual dimension (i.e., VF):
\begin{equation}\label{eq:fa_norm}
FA_i = \frac{F_{\text{CVLAT},i}}{F_{Q,i} + \varepsilon}.
\end{equation}} The overall FA Score is\deleted{ computed as} the mean of $FA_i$ across all questions\deleted{, yielding a value in the range [0, 1]}.

\noindent\textbf{Visual-Factual Reliance Index (VFRI)}\\
The VFRI combines the VF and FA scores into a single measure that reflects an LVLM's relative preference between visual information and factual knowledge. For each question $i$, the index is defined as:

\begin{equation}\label{eq:vfri}
\text{VFRI}_i = \frac{VF_i - FA_i}{VF_i + FA_i + \varepsilon}
\end{equation}
where $\varepsilon$ is the same small constant defined for Eq.~\ref{eq:vf_norm}\replaced{. The overall VFRI is the mean across all questions, ranging from $-1$ (strong factual preference) to $+1$ (strong visual preference).}{ when both scores are zero. The overall VFRI is computed as the mean across all questions:

\begin{equation}
\text{VFRI} = \frac{1}{N} \sum_{i=1}^{N} \text{VFRI}_i
\end{equation}

By definition, VFRI ranges from $-1$ to 1 and is interpreted as follows:
\begin{itemize}
    \item VFRI $\approx 1$: Strong preference for visual information over factual knowledge
    \item VFRI $\approx 0$: Equal weighting of both information sources
    \item VFRI $\approx -1$: Strong preference for factual knowledge over visual information
\end{itemize}}

\subsection{Experimental Results}\label{sec:exp2_results}

\added{\noindent\textbf{Human baseline.} Following the original VLAT study's correction-for-guessing convention (the visually-correct option as the scoring target), mean human accuracy on CVLAT is $53.71\%$ (SD $14.92$), and raw uncorrected accuracy is $60.76\%$ (SD $11.70$). The corrected score is statistically equivalent to both the original VLAT human baseline ($51.91\%$, SD $16.57$, $N=191$~\cite{lee2016vlat}) and to a recent Prolific VLAT replication reported in the Mini-VLAT study ($53.08\%$, SD $18.96$, $N=199$~\cite{pandey2023mini}) within a $\pm 10$ pp equivalence margin (see Appendix~C for the scoring formula and  formal equivalence test). We report this as a calibration result establishing that CVLAT is not substantially harder for humans than VLAT, rather than as a full revalidation of VLAT-equivalent difficulty. Applying the same formula with the \emph{Factual}-correct option as the target (i.e., counting how often participants chose the option matching real-world fact but contradicting the chart) yields a corrected score of $-9.46\%$ (SD $8.97$, unclamped), which is substantially below chance. In the aggregate, participants did not override the chart with prior factual knowledge but predominantly followed the visual representation, a behavior shared by all 30 participants individually. For this factual-scored diagnostic, we intentionally use the unclamped correction-for-guessing score to effectively detect below-chance factual selection. Conversely, the LVLM VF/FA normalization continues to employ the clamped score in Eq.~\ref{eq:score}.}

\begin{figure}
    \centering
    \includegraphics[width=\columnwidth]{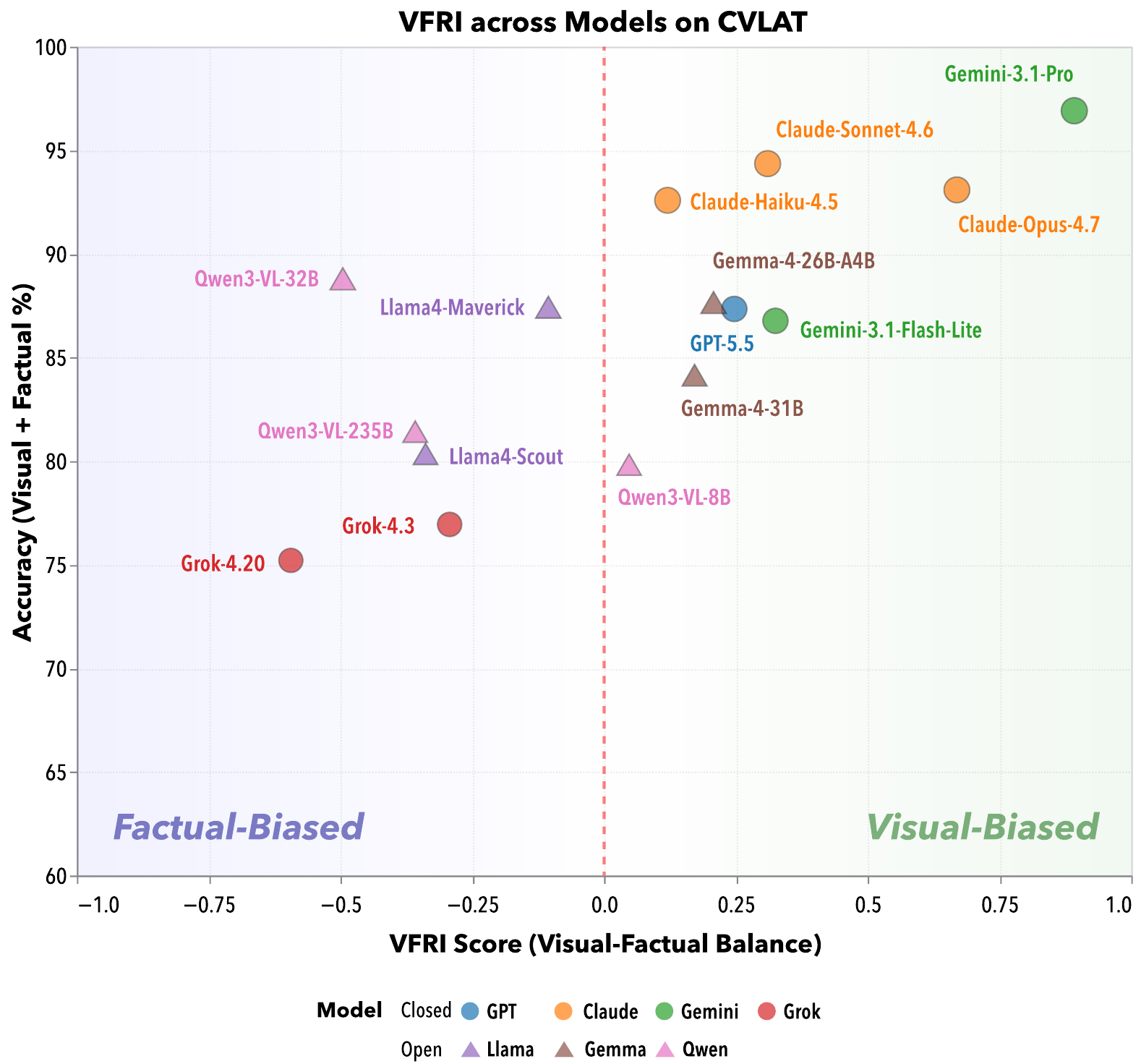}
    \caption{Relationship between Visual-Factual Reliance Index (VFRI) and accuracy. The x-axis shows VFRI scores ranging from $-1$ (strong factual preference) to 1 (strong visual preference), while the y-axis represents accuracy (percentage of visual + factual correct responses). Marker shapes indicate model source type (circles for closed-source, triangles for open-source), and colors represent different model families. \deleted{Note how models with similar VFRI scores (e.g., Gemma3-4b and Gemini-2.5-Flash near VFRI = $-0.1$) can have dramatically different accuracy levels, highlighting the importance of considering false response rates when evaluating model performance.} \added{Per-model 95\% bootstrap CIs are reported in Table~\ref{tab:cvlat_results}.}}
    \label{fig:vfri_vs_accuracy_baseline}
\end{figure}

\noindent\textbf{LVLM results.} Table~\ref{tab:cvlat_results} presents the comprehensive evaluation results of \replaced{15}{12} LVLMs on CVLAT, revealing \replaced{distinct prioritization patterns}{distinct patterns in how models prioritize between visual information and factual knowledge under conflicts}. \added{Figure~\ref{fig:vfri_vs_accuracy_baseline} provides a visual landscape of the corresponding VFRI--accuracy relationships across the suite.} \replaced{Based on the sign of their point-estimate VFRI, models are descriptively categorized into two distinct cohorts:}{Based on these results, the models can be broadly grouped into two primary categories according to their information prioritization tendencies:}

\begin{enumerate}
\item \textbf{Factual knowledge-oriented models}: \replaced{Grok-4.20, Qwen3-VL-32B, Qwen3-VL-235B, Llama4-Scout, Grok-4.3, and Llama4-Maverick}{GPT-4o, GPT-4.1, Claude-3.7-Sonnet, Grok-2-Vision, Gemma3-27b, Llama4-Maverick, Llama4-Scout, Qwen2.5-VL-72b, Gemini-2.5-Flash, and Gemma3-4b}. These models tend to favor factual knowledge over visual information when faced with counterfactual visualizations, although the strength of this tendency varies across architectures.

\item \textbf{Visualization-oriented models}: \replaced{Qwen3-VL-8B, Claude-Haiku-4.5, Gemma-4-31B, Gemma-4-26B-A4B, GPT-5.5, Claude-Sonnet-4.6, Gemini-3.1-Flash-Lite, Claude-Opus-4.7, and Gemini-3.1-Pro}{Gemini-2.5-Pro and Qwen2.5-VL-7b}. These models tend to follow visual encodings over factual knowledge when faced with counterfactual visualizations, although the strength of this tendency varies across architectures.
\end{enumerate}

\added{Nine of fifteen models have positive point-estimate VFRI. The most recent proprietary releases, including Claude-Opus-4.7 (VFRI $+0.669$), Gemini-3.1-Pro ($+0.892$), Claude-Sonnet-4.6 ($+0.310$), Gemini-3.1-Flash-Lite ($+0.325$), and GPT-5.5 ($+0.247$), fall on the visualization-oriented side, while both Meta Llama4 checkpoints and both xAI Grok checkpoints remain factual knowledge-oriented. While prior work~\cite{hong2025llms} reported that LVLMs rely heavily on pre-existing knowledge when interpreting visualizations, our results show that this tendency is not uniform across architectures or generations. Individual LVLMs exhibit distinct and measurable prioritization patterns rather than a single shared bias.}

\added{At the strong-visualization end, Gemini-3.1-Pro achieves the highest VFRI ($+0.892$) and a near-ceiling VF Score (0.987) while maintaining a remarkably low false-response rate (3.09\%), indicating robust visual interpretation even when visualizations present counterfactual information. Claude-Opus-4.7 follows closely (VFRI $+0.669$, VF Score 0.792, false rate 6.91\%). Both models also have near-ceiling chart-reading capability references (Anon V $\geq 91\%$), so their positive VFRI reflects an actual visual preference rather than a capability asymmetry. At the opposite end, Grok-4.20 has the strongest negative VFRI ($-0.594$). Even after capability normalization, the model selects the factual option much more often than the visual option, consistent with aggregate factual-prior reliance relative to its measured chart-reading capability (Anon V $35.83\%$).}

\added{A notable contrast emerges within the Qwen3-VL family. The 8B checkpoint has a weak positive, near-boundary VFRI ($+0.048$), whereas the 32B and 235B checkpoints show clear factual orientation (VFRI $-0.496$ and $-0.358$, respectively). Although recent research suggests that larger models can store substantially more factual knowledge, with capacity scaling linearly with model size~\cite{allen2024physics, lu2024scaling}, our Q-only probe shows that factual-prior availability is nearly identical across the three sizes ($62.73\%$, $62.81\%$, and $62.90\%$ for 8B, 32B, and 235B). The within-family contrast therefore more directly reflects differences in how strongly each checkpoint arbitrates in favor of factual priors when those priors are equally available.}

\added{This pattern does not generalize across families. Both Llama4 checkpoints lean factual (Maverick $-0.106$, Scout $-0.339$), both Gemma-4 checkpoints lean visual (positive VFRI), and all three Claude checkpoints lean visual. The Llama4 lean is notable because both checkpoints possess reasonable chart-reading capability (Anon V $72.03$ and $60.69$) yet exhibit Q-only factual rates ($84.83$ and $76.65$) comparable to the strongest factual-prior models in the suite. Notably, recent proprietary frontier models, including Gemini-3.1-Pro ($+0.892$), Claude-Opus-4.7 ($+0.669$), and Claude-Sonnet-4.6 ($+0.310$), remain visualization-oriented despite their undisclosed but presumably larger parameter counts, suggesting that scale is only one factor among several---such as training-data composition, RLHF objectives, and instruction-tuning recipes---that shape arbitration behavior. Because vision-encoder, connector, and tokenizer details are not disclosed for the proprietary endpoints in our suite, systematically disentangling the specific impact of architecture versus model family remains a target for future work.}

\added{Relating these patterns to the human baseline reported above sharpens the human--model comparison. Humans are visualization-oriented without exception (all 30 participants have positive VFRI, with factual-scored accuracy well below chance), whereas the models split into oriented groups. The human--model divergence is therefore concentrated in the factual knowledge-oriented checkpoints (Grok-4.20, Grok-4.3, Qwen3-VL-32B, Qwen3-VL-235B, and Llama4-Scout), which select the factually-correct, chart-contradicting option far more often than any human participant did, whereas the visualization-oriented models resolve the conflict much as the human cohort does.}

\section{Experiment Three: Steering LVLMs' Information Prioritization Through Prompt Engineering}
\label{sec:exp3}

In our third experiment, we investigate \textbf{RQ3}: To what extent can prompt engineering shift LVLMs' information prioritization between visual information and factual knowledge when the two sources conflict?

This investigation aims to determine whether the information prioritization patterns identified in Experiment 2 can be influenced through explicit prompt engineering, or whether they instead reflect more stable model-specific tendencies.\deleted{ Specifically, we examine whether explicit prompting can: (i)~redirect visualization-oriented models (Gemini-2.5-Pro, Qwen2.5-VL-7b) toward prioritizing factual knowledge; (ii)~redirect factual-knowledge-oriented models (ten models including GPT-4.1 and Claude-3.7-Sonnet) toward prioritizing visual information; (iii)~distinguish models whose prioritization is responsive to prompting from those whose behavior remains largely unchanged, indicating stronger intrinsic tendencies.} \replaced{Every model in our suite receives both the factual-priority and the visual-priority prompts, regardless of baseline orientation.}{Building on the two-group distinction established in Experiment 2---factual-knowledge-oriented (ten models) and visualization-oriented (two models)---we select models from both groups to evaluate the effectiveness of prompt-based interventions. To focus on models with clearly expressed preferences, we include the two visualization-oriented models and eight factual-knowledge-oriented models exhibiting the strongest biases, as measured by the absolute value of VFRI. We exclude Gemini-2.5-Flash and Gemma3-4b, whose $|$VFRI$|$ values fall below 0.15. Consistent with established effect size guidelines, values below approximately $0.10$--$0.15$ are commonly interpreted as negligible or ``very small''~\cite{gignac2016effect, funder2019evaluating}. Accordingly, we interpret these low $|$VFRI$|$ values as indicating no clear or stable information prioritization preference.}

\subsection{Prompt Design}

We adopt an explicit prompting strategy to investigate whether prompt engineering can shift models' information prioritization preferences (see Appendix~A for full prompt texts). Two contrasting prompts are designed: the \textbf{factual-priority prompt} instructs models to prioritize their factual knowledge over visual information when contradictions arise, while the \textbf{visual-priority prompt} instructs models to treat visual data as ground truth and respond based on visual information, even when it conflicts with their factual knowledge.

\subsection{Experimental Protocol and Analysis Methodology}
We follow the same experimental protocol as Experiment 2, using CVLAT with 48 questions and 120 permutations per question.
\replaced{Every model is evaluated under both the factual-priority and visual-priority prompts, yielding 11{,}520 trials per model and 172{,}800 trials across all 15 models.}{Visualization-oriented models are evaluated under the factual priority prompt, whereas factual-knowledge-oriented models are evaluated under the visual priority prompt.
This design results in 5,760 trials per model, totaling 57,600 trials across ten models.}

For analysis, we compute the VF Score, FA Score, and VFRI under the prompt-engineering condition and compare them with the corresponding baseline scores from Experiment 2.
Prompt effectiveness is quantified as the change ($\Delta$) between intervention and baseline scores, where positive $\Delta$VFRI values indicate shifts toward prioritizing visual information and negative values indicate shifts toward prioritizing factual knowledge.
Statistical significance is assessed using paired bootstrap testing (10,000 iterations, $\alpha$ = 0.05), with 95\% confidence intervals and significance labels reported in Table~\ref{tab:prompt_engineering_results}.

\begin{table*}[ht]
\centering
\caption{Effects of prompt engineering on LVLM information prioriti\replaced{zation, grouped by controllability profile (Sec.~\ref{sec:exp3} narrative). All values are Visual-Factual Reliance Index (VFRI). The two $\Delta$VFRI columns flank the baseline condition ($\Delta$ = Prompted $-$ Baseline, where positive shifts indicate a movement toward visual evidence and negative shifts toward factual priors). Brackets below each $\Delta$ value report the 95\% confidence interval from a paired bootstrap ($B = 10{,}000$, shift method). The Trajectory column shows the three VFRI positions on a $[-1, +1]$ axis, color-coded for interpretation: red for factual-priority, black for baseline, and blue for visual-priority. Significance levels: *** $p \leq 0.001$, ** $p \leq 0.01$, * $p \leq 0.05$, and $ns$ for not significant}{sation. Visualization-oriented models received factual-priority prompts, while factual-knowledge-oriented models received visual-priority prompts. Significance levels: *** $p \leq 0.001$, ** $p \leq 0.01$, * $p \leq 0.05$, $ns$ for not significant (paired bootstrap, $B = 10{,}000$)}.}
\label{tab:prompt_engineering_results}
\newcommand{\dcell}[2]{\makebox[0.75cm][r]{$#1$}\hspace{0.35cm}\makebox[0.4cm][l]{#2}}
\newcommand{\traj}[1]{\includegraphics[width=2.4cm]{figs/trellis_thumbnails/#1.pdf}}
\begin{tabular}{l|c|c|c|c|c|c}
\toprule
\textbf{Model} & \added{\textbf{Factual Prompted}} & \added{\textbf{$\Delta_F$\,VFRI}} & \textbf{Baseline} & \added{\textbf{$\Delta_V$\,VFRI}} & \added{\textbf{Visual Prompted}} & \added{\textbf{Trajectory}} \\
\midrule
\multicolumn{7}{l}{\added{\textit{Symmetric responders}}} \\
\midrule
\multirow{2}{*}{Claude-Haiku-4.5$^{\dagger}$} & \multirow{2}{*}{$+0.055$} & \dcell{-0.066}{ns} & \multirow{2}{*}{$+0.120$} & \dcell{+0.177}{ns} & \multirow{2}{*}{$+0.297$} & \multirow{2}{*}{\traj{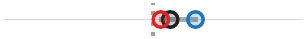}} \\
 & & {\scriptsize $[-0.221, +0.080]$} & & {\scriptsize $[-0.013, +0.381]$} & & \\
\midrule
\multirow{2}{*}{Gemma-4-31B} & \multirow{2}{*}{$-0.268$} & \dcell{-0.440}{*} & \multirow{2}{*}{$+0.172$} & \dcell{+0.612}{***} & \multirow{2}{*}{$+0.784$} & \multirow{2}{*}{\traj{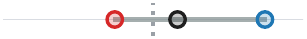}} \\
 & & {\scriptsize $[-0.794, -0.064]$} & & {\scriptsize $[+0.377, +0.862]$} & & \\
\midrule
\multirow{2}{*}{Gemma-4-26B-A4B} & \multirow{2}{*}{$-0.081$} & \dcell{-0.289}{*} & \multirow{2}{*}{$+0.208$} & \dcell{+0.340}{***} & \multirow{2}{*}{$+0.548$} & \multirow{2}{*}{\traj{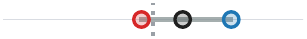}} \\
 & & {\scriptsize $[-0.554, -0.015]$} & & {\scriptsize $[+0.189, +0.513]$} & & \\
\midrule
\multirow{2}{*}{Claude-Sonnet-4.6} & \multirow{2}{*}{$-0.354$} & \dcell{-0.665}{***} & \multirow{2}{*}{$+0.310$} & \dcell{+0.095}{*} & \multirow{2}{*}{$+0.405$} & \multirow{2}{*}{\traj{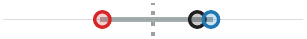}} \\
 & & {\scriptsize $[-0.903, -0.433]$} & & {\scriptsize $[+0.012, +0.208]$} & & \\
\midrule
\multirow{2}{*}{Gemini-3.1-Flash-Lite} & \multirow{2}{*}{$-0.687$} & \dcell{-1.012}{***} & \multirow{2}{*}{$+0.325$} & \dcell{+0.386}{**} & \multirow{2}{*}{$+0.711$} & \multirow{2}{*}{\traj{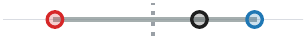}} \\
 & & {\scriptsize $[-1.286, -0.735]$} & & {\scriptsize $[+0.191, +0.612]$} & & \\
\midrule
\multirow{2}{*}{Claude-Opus-4.7} & \multirow{2}{*}{$-0.497$} & \dcell{-1.166}{***} & \multirow{2}{*}{$+0.669$} & \dcell{+0.241}{**} & \multirow{2}{*}{$+0.910$} & \multirow{2}{*}{\traj{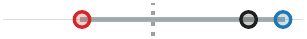}} \\
 & & {\scriptsize $[-1.438, -0.892]$} & & {\scriptsize $[+0.093, +0.415]$} & & \\
\midrule
\multicolumn{7}{l}{\added{\textit{F-priority-collapsing}}} \\
\midrule
\multirow{2}{*}{Qwen3-VL-32B} & \multirow{2}{*}{$+0.264$} & \dcell{+0.759}{***} & \multirow{2}{*}{$-0.496$} & \dcell{+1.096}{***} & \multirow{2}{*}{$+0.600$} & \multirow{2}{*}{\traj{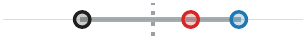}} \\
 & & {\scriptsize $[+0.507, +1.016]$} & & {\scriptsize $[+0.819, +1.369]$} & & \\
\midrule
\multirow{2}{*}{Qwen3-VL-235B} & \multirow{2}{*}{$-0.036$} & \dcell{+0.322}{*} & \multirow{2}{*}{$-0.358$} & \dcell{+0.733}{***} & \multirow{2}{*}{$+0.375$} & \multirow{2}{*}{\traj{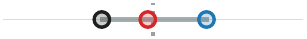}} \\
 & & {\scriptsize $[+0.049, +0.614]$} & & {\scriptsize $[+0.446, +1.020]$} & & \\
\midrule
\multicolumn{7}{l}{\added{\textit{V-priority-insensitive}}} \\
\midrule
\multirow{2}{*}{GPT-5.5} & \multirow{2}{*}{$-0.537$} & \dcell{-0.784}{***} & \multirow{2}{*}{$+0.247$} & \dcell{-0.062}{ns} & \multirow{2}{*}{$+0.185$} & \multirow{2}{*}{\traj{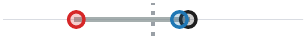}} \\
 & & {\scriptsize $[-1.011, -0.555]$} & & {\scriptsize $[-0.184, +0.039]$} & & \\
\midrule
\multirow{2}{*}{Gemini-3.1-Pro$^{\ddagger}$} & \multirow{2}{*}{$-0.839$} & \dcell{-1.731}{***} & \multirow{2}{*}{$+0.892$} & \dcell{+0.028}{ns} & \multirow{2}{*}{$+0.920$} & \multirow{2}{*}{\traj{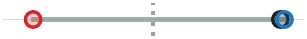}} \\
 & & {\scriptsize $[-1.883, -1.548]$} & & {\scriptsize $[-0.002, +0.078]$} & & \\
\midrule
\multicolumn{7}{l}{\added{\textit{F-priority-insensitive}}} \\
\midrule
\multirow{2}{*}{Grok-4.20} & \multirow{2}{*}{$-0.473$} & \dcell{+0.121}{ns} & \multirow{2}{*}{$-0.594$} & \dcell{+0.332}{***} & \multirow{2}{*}{$-0.262$} & \multirow{2}{*}{\traj{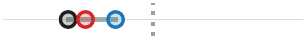}} \\
 & & {\scriptsize $[-0.080, +0.358]$} & & {\scriptsize $[+0.149, +0.535]$} & & \\
\midrule
\multirow{2}{*}{Llama4-Scout} & \multirow{2}{*}{$-0.524$} & \dcell{-0.186}{ns} & \multirow{2}{*}{$-0.339$} & \dcell{+0.416}{***} & \multirow{2}{*}{$+0.077$} & \multirow{2}{*}{\traj{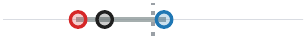}} \\
 & & {\scriptsize $[-0.417, +0.038]$} & & {\scriptsize $[+0.187, +0.651]$} & & \\
\midrule
\multirow{2}{*}{Grok-4.3} & \multirow{2}{*}{$-0.263$} & \dcell{+0.030}{ns} & \multirow{2}{*}{$-0.293$} & \dcell{+0.166}{**} & \multirow{2}{*}{$-0.127$} & \multirow{2}{*}{\traj{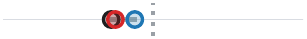}} \\
 & & {\scriptsize $[-0.053, +0.119]$} & & {\scriptsize $[+0.050, +0.288]$} & & \\
\midrule
\multirow{2}{*}{Llama4-Maverick} & \multirow{2}{*}{$-0.107$} & \dcell{-0.001}{ns} & \multirow{2}{*}{$-0.106$} & \dcell{+0.340}{*} & \multirow{2}{*}{$+0.234$} & \multirow{2}{*}{\traj{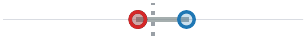}} \\
 & & {\scriptsize $[-0.246, +0.251]$} & & {\scriptsize $[+0.062, +0.625]$} & & \\
\midrule
\multirow{2}{*}{Qwen3-VL-8B} & \multirow{2}{*}{$+0.107$} & \dcell{+0.060}{ns} & \multirow{2}{*}{$+0.048$} & \dcell{+0.278}{*} & \multirow{2}{*}{$+0.326$} & \multirow{2}{*}{\traj{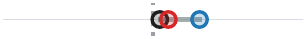}} \\
 & & {\scriptsize $[-0.235, +0.363]$} & & {\scriptsize $[+0.024, +0.553]$} & & \\
\bottomrule
\end{tabular}
\begin{tablenotes}
\fontsize{7pt}{7pt}\selectfont
\item * All VFRI values are capability-normalized per Sec.~\ref{sec:exp2}; baseline column matches Table~\ref{tab:cvlat_results} exactly.
\item $^{\dagger}$ Both shifts are directionally consistent with the Symmetric profile but neither reaches statistical significance, indicating an underpowered observation rather than confirmed bidirectional control.
\item $^{\ddagger}$ Near-ceiling baseline (VFRI $= +0.892$); the absence of a significant $\Delta_V$ reflects limited headroom for further visual gain rather than insensitivity to the visual-priority prompt.
\end{tablenotes}
\end{table*}

\subsection{Experimental Results}

\begin{figure}[tb]
    \centering
    \includegraphics[width=\columnwidth]{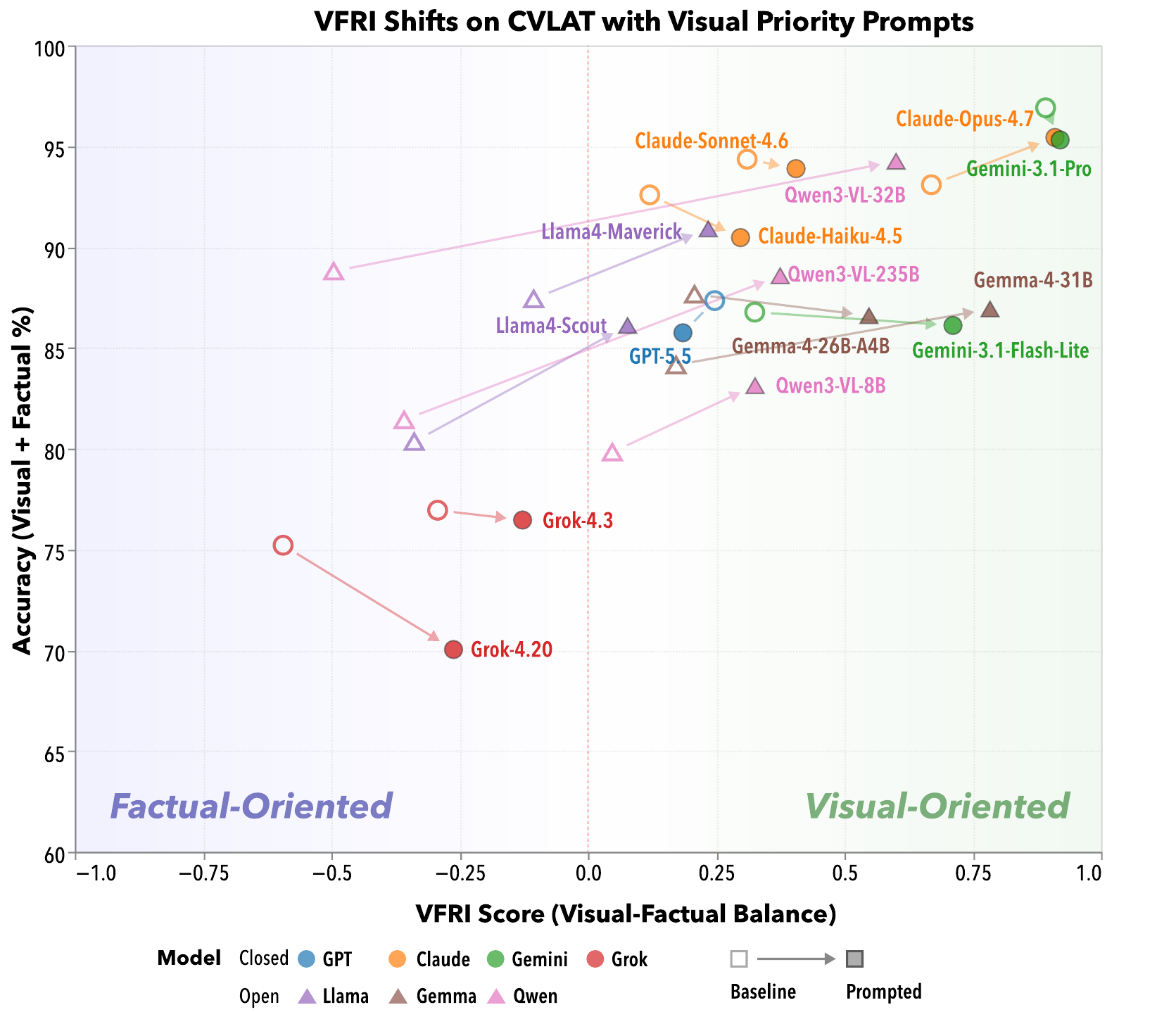}
    \caption{\added{Visual-priority prompt effect. Each model contributes a hollow baseline marker (start) and a filled visual-priority marker (end) connected by an arrow indicating the direction of change in VFRI and accuracy. Rightward movement along the VFRI axis indicates the prompt successfully pulled the model toward visual evidence. Per-model 95\% paired-bootstrap CIs for $\Delta_V$ VFRI are reported in Table~\ref{tab:prompt_engineering_results}.}}
    \label{fig:nvfri_v_2point}
\end{figure}

\begin{figure}[tb]
    \centering
    \includegraphics[width=\columnwidth]{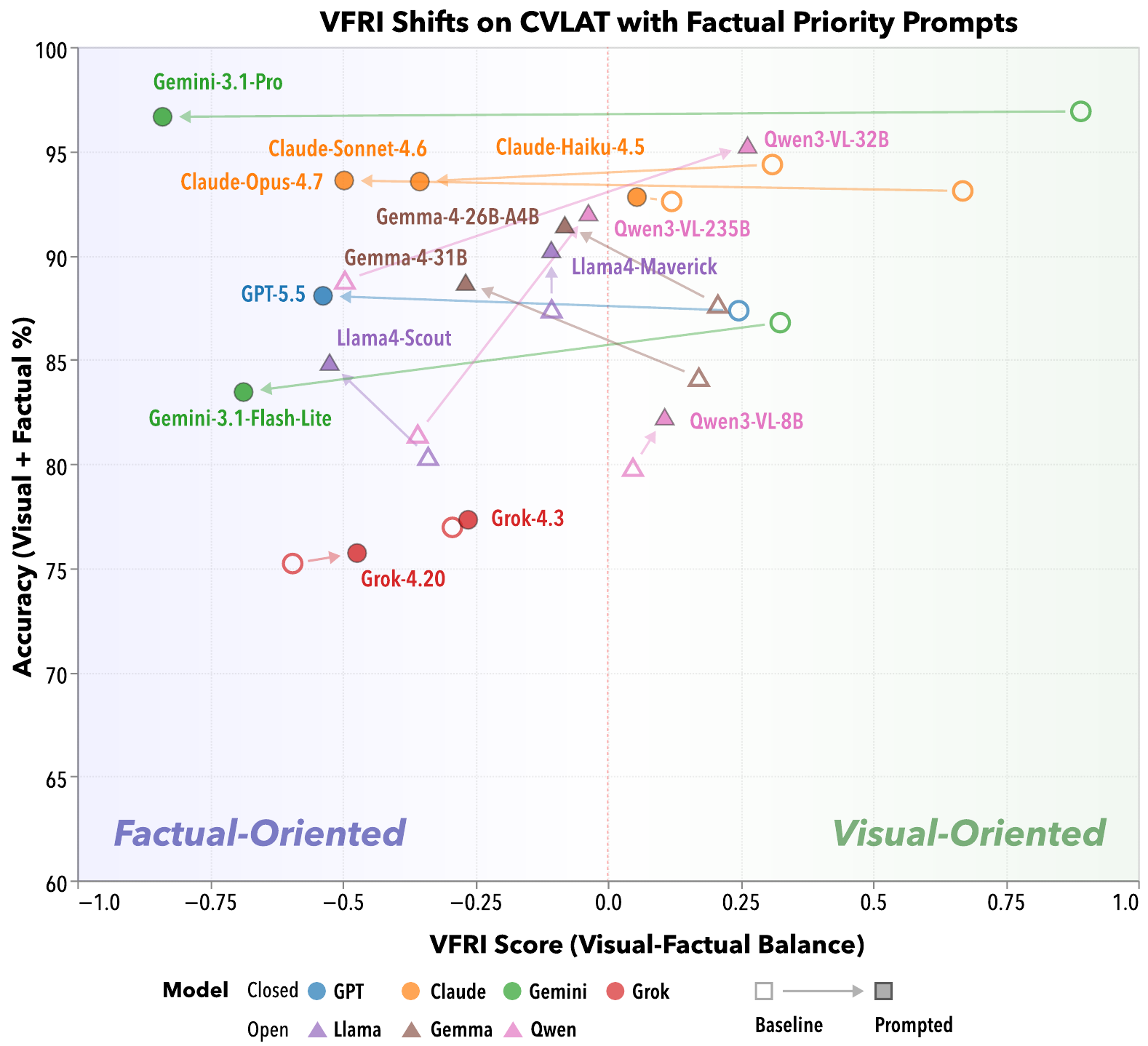}
    \caption{\added{Factual-priority prompt effect. Each model contributes a hollow baseline marker (start) and a filled factual-priority marker (end) connected by an arrow indicating the direction of change in VFRI and accuracy. Leftward movement indicates the prompt pulled the model toward factual priors, while rightward movement is the F-priority-collapsing anomaly discussed in Sec.~\ref{sec:exp3} and Sec.~\ref{sec:discussion}. Per-model 95\% paired-bootstrap CIs for $\Delta_F$ VFRI are reported in Table~\ref{tab:prompt_engineering_results}.}}
    \label{fig:nvfri_f_2point}
\end{figure}

\added{Table~\ref{tab:prompt_engineering_results} and Figures~\ref{fig:nvfri_v_2point}--\ref{fig:nvfri_f_2point} summarize the effects of prompt engineering on LVLMs' information prioritization. Under the full-factorial design in which every model receives both factual-priority and visual-priority prompts, models fall into four distinct controllability profiles, which serve as the organizational basis for the rows of Table~\ref{tab:prompt_engineering_results}.}

\added{\noindent\textbf{Symmetric responders.} Claude-Opus-4.7 exhibited one of the most pronounced responses to prompt engineering, shifting from strongly visualization-oriented under the baseline condition (VFRI $+0.669$) to clearly factual knowledge-oriented stance under the factual-priority prompt (VFRI $-0.497$, $\Delta_F = -1.166$, $p < 0.001$), while extending further toward visual reliance under the visual-priority prompt (VFRI $+0.910$, $\Delta_V = +0.241$, $p < 0.01$). Gemini-3.1-Flash-Lite shows a comparably broad bidirectional range ($\Delta_F = -1.012$, $\Delta_V = +0.386$, both $p \leq 0.01$). Claude-Sonnet-4.6, Gemma-4-31B, and Gemma-4-26B-A4B also belong to this profile, shifting significantly in both directions as instructed. Claude-Haiku-4.5 is directionally consistent in both directions but neither shift reaches statistical significance, suggesting an underpowered variant of the same underlying pattern.}

\added{\noindent\textbf{F-priority-collapsing models.} Qwen3-VL-32B and Qwen3-VL-235B exhibited a counter-intuitive pattern: the factual-priority prompt triggered a shift \emph{toward} visual processing instead of deepening further factual alignment (positive and significant $\Delta_F$: $+0.759$ *** for 32B; $+0.322$ * for 235B), while their visual-priority shifts remained uniformly positive and substantial ($\Delta_V$ up to $+1.096$ for 32B). The pattern is accompanied by a sharp surge in generation length: median completion tokens expanded from 7 in the baseline to 212 under the factual-priority condition for Qwen3-VL-32B, and from 3 to 136 for Qwen3-VL-235B. Rather than providing a direct answer, factual-priority outputs typically begin with explicit breakdown of the chart. On a stacked-area GDP item, for example, Qwen3-VL-32B transitioned from a single-token baseline response of ``(a)'' to an extended text starting, ``\emph{The chart provided shows GDP Growth Trends for the US, China, and Japan from 2009 to 2014 \ldots\ In 2009, the US GDP is around \$14{,}000 billion \ldots}''. We interpret this in Sec.~\ref{sec:discussion} as a deliberation-activation effect. By compelling the model to verify facts, the factual-priority prompt elicits longer, chart-referencing outputs that appear to surface visual evidence the terse baseline elided, eventually driving the model's final state away from (not toward) the prompted direction.}

\added{\noindent\textbf{V-priority-insensitive models.} GPT-5.5 exhibited a strong factual-priority response ($\Delta_F = -0.784$, $p < 0.001$) but no detectable visual-priority response ($\Delta_V = -0.062$, ns); uniquely within the evaluation suite, its visual-priority point estimate was mildly negative. This asymmetry persists despite the model possessing an intact chart-reading capability. For example, on the tech-company quarterly-revenue bar chart (evaluating the comparison \emph{``IBM Q2 revenue is larger than Facebook's''}), GPT-5.5 achieves a visual-correct rate of $100\%$ on the anonymized chart and $97\%$ at baseline, yet this rate drops sharply to $29\%$ under the visual-priority prompt. This indicates that chart-reading capability does not translate into visual reliance under conflict for this model, which instead defaults to its factual prior. Gemini-3.1-Pro exhibits a comparable profile, acting as a near-ceiling variant. Because it operates with a high baseline VFRI of $+0.892$, there is essentially no remaining headroom for further visual gain, resulting in a stark shift under the factual prompt but stagnation under the visual one ($\Delta_F = -1.731$ ***, $\Delta_V = +0.028$ ns).}

\added{\noindent\textbf{F-priority-insensitive models.} The Grok and Llama4 families, along with Qwen3-VL-8B, exhibited the mirror-image pattern: they demonstrated significant visual-priority responses ($\Delta_V$ ranging from $+0.166$ to $+0.416$, all $p \leq 0.05$; see Table~\ref{tab:prompt_engineering_results}) but factual-priority effects that fell below significance. These checkpoints comply effectively with the visual-priority instruction, yet the factual-priority instruction produces no statistically detectable shift toward factual reliance.}
\smallskip

\added{Taken together, prompt engineering is a viable but non-universal strategy for steering information prioritization. The heterogeneity across models is substantial, including divergent responses between comparably capable proprietary models (e.g., GPT-5.5 vs.\ Claude-Opus-4.7). This controllability is often non-reciprocal; single-direction insensitivity manifests in both visual and factual directions, and the F-priority-collapse profile reveals instances where factual-priority prompts inadvertently trigger the inverse behavior. Importantly, standard visualization-literacy benchmarks fail to predict these behavioral profiles. Higher-performing Gemini-3.1-Pro (top VLAT-Normal and VLAT-Explain in the suite) and GPT-5.5 (third-highest VLAT-Explain at 93.85\%) are classified as V-priority-insensitive, while lower-performing Llama4 variants exhibit F-priority-insensitive. Within-family controllability patterns are mixed. While the Claude, Gemma-4, Llama4, and Grok families consistently fall within a single profile, Qwen3-VL and Gemini diverge internally. The Qwen3-VL lineage shows a clear scale-associated split (8B in F-priority-insensitive, 32B and 235B in F-priority-collapsing) that mirrors the baseline VFRI gradient analyzed in Sec.~\ref{sec:exp2}. Conversely, the Gemini divergence occurs across distinct scale tiers (Pro in V-priority-insensitive, Flash-Lite in Symmetric). The scale-dependent split observed in Qwen3-VL suggests that a model's intrinsic arbitration tendency and its susceptibility to prompt-based scaffolding covary with scale within the Qwen3-VL family.}

\section{Discussion}
\label{sec:discussion}

Recent studies have reported strong performance of LVLMs on visualization interpretation tasks, raising an important question: to what extent does such performance reflect genuine visualization literacy, as opposed to reliance on pre-trained factual knowledge? Our results \replaced{show}{provide empirical evidence, grounded in systematic evaluation and measurable performance metrics,} that accuracy alone is insufficient to answer this question.
By disentangling visual correctness from factual correctness, we show that \replaced{high benchmark scores alone do not reveal how models arbitrate under conflict. Although nine of 15 models show positive point-estimate VFRI under the baseline CVLAT condition, Experiment 3 demonstrates that this visual orientation is by no means uniformly stable under prompt-based interventions. Rather, these intervention effects remain heavily constrained, direction-asymmetric, and highly model-dependent. Notably, this behavioral divergence has important implications for how visualization literacy should be evaluated and interpreted in LVLM-based analytical systems.}{many LVLMs achieve high benchmark scores by prioritizing factual knowledge, whereas only a small subset consistently follows visual evidence under conflict. This distinction has important implications for how visualization literacy should be evaluated and interpreted in LVLM-based systems.}

In line with prior empirical evaluations of LVLM visualization literacy, we assess a \replaced{state-of-the-art LVLM suite on established benchmarks (Sec.~\ref{sec:exp1}, Table~\ref{tab:results_experiment1}). Several models approach or exceed the 65.5\% human VLAT baseline yet generally perform worse on reVLAT.}{broad set of state-of-the-art LVLMs using established benchmarks. On VLAT, two open-source LVLMs (Llama4-Maverick and Qwen2.5-VL-72b) achieve performance approaching human baselines, while three proprietary LVLMs (GPT-4.1, Gemini-2.5-Pro-Preview, and Gemini-2.5-Flash-Preview) exceed human performance under the Normal prompt condition. In contrast, performance on reVLAT, which uses randomized data, drops substantially. Assuming a consistent human baseline across VLAT and reVLAT, only Gemini-2.5-Pro-Preview continues to outperform humans under the Normal prompt condition.}
This divergence between VLAT and reVLAT performance underscores a core limitation of accuracy-based evaluation: high scores on aligned benchmarks do not necessarily indicate robust visualization interpretation. Instead, they may reflect reliance on factual priors that break down when visual and factual signals diverge. This observation directly motivates our disentanglement framework, which exposes systematically different response patterns, consistent with visual interpretation versus factual recall, that remain indistinguishable under conventional evaluation protocols.

Interpreted through our proposed quadrant framework, these results yield two key insights. First, on benchmarks such as VLAT where visual encodings align with real-world factual knowledge, high accuracy may be driven by recall of pre-trained knowledge rather than interpreting visual information, leading to inflated estimates of visualization literacy. Second, on benchmarks with \replaced{randomized}{counterfactual} data, such as reVLAT, models' tendency to prioritize factual knowledge over visual evidence can suppress correct visual interpretation, resulting in underestimation of their actual visualization literacy.
These complementary biases help explain the inconsistent findings in prior studies \cite{bendeck2024empirical, hong2025llms}. Differences in benchmark design cause evaluations to capture varying mixtures of visualization interpretation and factual recall, leading to inconsistent assessments of the same models even when similar performance metrics are used.

Building on this insight, we introduce CVLAT to explicitly investigate which information LVLMs prioritize when visual information and factual knowledge (i.e., pre-trained knowledge) conflict, and define metrics to quantify their tendencies. Our results reveal two distinct model groups (factual knowledge-oriented and visualization-oriented)\replaced{. Whereas Hong et al.~\cite{hong2025llms} document that LVLMs rely on prior knowledge as a population-level pattern, our capability-normalized CVLAT metrics show that this reliance is not uniform: individual models exhibit distinct, measurable, and partially steerable prioritization patterns.}{---highlighting systematic differences in information prioritization. This finding extends previous observations that LVLMs often rely on factual knowledge when solving problems \cite{hong2025llms} by showing such reliance is not uniform: individual LVLMs exhibit distinct and measurable prioritization patterns rather than a single, shared bias.} \added{CVLAT differs from reVLAT in that it preserves the factual signal and forces the two into conflict, surfacing arbitration behavior that is diagnostically relevant whenever visual evidence and prior knowledge disagree.}

Finally, we examine whether explicit prompting can modulate LVLMs' information prioritization\replaced{ (Sec.~\ref{sec:exp3}). Models exhibit four prompt-controllability profiles. Symmetric responders move in the expected direction under both factual-priority and visual-priority prompts. F-priority-collapsing responders unexpectedly move toward visual evidence when instructed to prioritize factual knowledge. The remaining two profiles are one-sided: V-priority-insensitive models respond only to factual-priority prompts, whereas F-priority-insensitive models respond only to visual-priority prompts. Single-direction prompt failure is therefore a recurring outcome rather than an isolated curiosity, and prompt-controllability cannot be assumed to be reciprocal.}{. We observe significant preference shifts in four models (Gemini-2.5-Pro, GPT-4.1, Llama4-Scout, and Llama4-Maverick), indicating that their prioritization strategies are at least partially responsive to prompt instructions. In contrast, Qwen2.5-VL-7b shows consistently strong visual reliance across prompting conditions, suggesting that its behavior may stem from the limited availability of factual priors rather than flexible or deliberate prioritization between information sources.}

\added{The output-length analysis reported in Sec.~\ref{sec:exp3} supports a deliberation-activation interpretation of the F-priority-collapse pattern (the factual-priority prompt activates chart engagement that the baseline behavior elided) rather than active override of visual evidence. We caution that this signature is suggestive rather than definitive. The factual-priority instruction is conditional (triggering only under recognized conflict) while the visual-priority instruction is unconditional, and sparse factual priors are an additional confound. Disentangling these mechanisms more precisely is left for future work.}

Overall, the three evaluation instruments (VLAT, reVLAT, and CVLAT) play complementary roles\replaced{. Together, they provide an assessment by isolating and operationalizing different capacities within LVLM visualization literacy. Specifically, VLAT measures perceptual decoding under aligned conditions, reVLAT measures perceptual decoding with factual priors largely neutralized via data randomization, and CVLAT captures arbitration dynamics when visual and factual signals conflict. The Q-only condition, administered alongside CVLAT as a capability reference, indexes factual grounding (i.e., factual-prior availability without visual evidence), and the capability-normalization applied in VF/FA incorporates this factual-grounding layer directly into the arbitration metric. This multi-instrument approach addresses diagnostic gaps that no single benchmark resolves. Consequently, CVLAT directly reveals how models prioritize conflicting sources.}{. VLAT provides baseline comparability with human performance but may conflate knowledge recall with visual interpretation when the two align. reVLAT isolates visual processing by removing factual alignment with randomized data, yet can underestimate visualization literacy for models that override visual information with factual knowledge. CVLAT uniquely reveals how models prioritize conflicting information sources, providing critical insights for deployment contexts in which faithful interpretation of visual evidence is essential.}

\added{Consistent with this design, our human cohort ($N=30$) under the same Normal prompt setting systematically followed the chart rather than oscillating between visual and factual options, with factual-scored accuracy well below chance (Sec.~\ref{sec:exp2_results}, Appendix~C). This robust human baseline indicates that the counterfactual construction within our benchmark does not collapse into an ambiguity-driven guessing game, even under minimal prompting.}

\added{Arbitration is central to deployed literacy because real-world workflows routinely encounter charts whose correct interpretation depends on which competing signal to trust. Therefore, CVLAT serves as a diagnostic designed to match a model's intrinsic orientation with specific use-case requirements, rather than functioning as a conventional, one-dimensional model leaderboard. Visual fidelity and factual prioritization represent descriptive orientations rather than normative judgments.}

\subsection{Implications for Visual Analytics System Design}

Our findings have practical implications for the deployment of LVLMs in visual analytics systems, particularly in settings where faithful interpretation of visual evidence is critical.

\textbf{Model Selection}: For applications requiring faithful visual interpretation (e.g., exploratory data analysis, anomaly detection, or visual validation), visualization-oriented models \replaced{(indicated by a high positive VFRI in Table~\ref{tab:cvlat_results})}{like Gemini-2.5-Pro} are preferable, as these models more reliably report what is visually encoded, even when the observed stimuli directly contradict their intrinsic factual expectations. \added{Conversely, in deployments where misleading or adversarial charts are a concern, factual knowledge-oriented models may be preferable, as their tendency to favor factual priors over conflicting visual encodings could mitigate certain forms of visual misinformation, though we did not directly evaluate adversarial charts.} For knowledge-intensive tasks where visualizations largely align with real-world facts, factual knowledge-oriented models may be more appropriate. \deleted{Importantly, practitioners should not interpret balanced VFRI values as evidence of balanced capability: intermediate VFRI values can arise from high false response rates, indicating instability or confusion rather than flexible information integration.}

\textbf{Prompt Engineering}: Our results demonstrate that prompt engineering has limited and model-dependent effectiveness in redirecting information prioritization. \added{While most evaluated models are responsive to at least one priority prompt, only approximately one-third exhibit bidirectional controllability, and some respond effectively in only a single direction (asymmetric sensitivity). Consequently, for systems requiring dynamic and adaptable information prioritization, developers should prefer models with empirically verified bidirectional controllability rather than assuming prompt engineering alone suffices.} \added{Importantly, high visualization-literacy benchmark scores do not necessarily imply prompt-controllability. GPT-5.5 attains 93.85\% on VLAT-Explain and a high anonymized-visual baseline (75.3\%) on CVLAT, yet shows no significant visual-priority response (Sec.~\ref{sec:exp3}). Visual capability and prompt-controllability should therefore be verified separately rather than inferred from one another.}

\textbf{Mitigation Strategies}: \replaced{For high-stakes applications, robustness can be improved by pairing models with contrasting orientations, incorporating CVLAT-style uncertainty estimation, and adding human-in-the-loop review under strong visual--factual conflict or low model confidence.}{For high-stakes or safety-critical applications, we recommend several complementary mitigation strategies: (1) employing multiple LVLMs with contrasting prioritization tendencies (e.g., pairing visualization-oriented and factual knowledge-oriented models) to cross-validate interpretations; (2) incorporating uncertainty estimation informed by CVLAT-style evaluations during deployment; (3) integrating human-in-the-loop workflows for cases where models encounter strong visual-factual conflicts or exhibit low confidence.}

\subsection{Limitations and Future Work}

While our study provides valuable insights, several limitations warrant future investigation.

\textbf{Scope Limitations}. Our evaluation covers common visualization types such as bar charts, line graphs, scatterplots, and pie charts\replaced{. Generalizing our findings to more intricate chart types and formats (e.g., network diagrams, heatmaps with complex encoding schemes, 3D plots, domain-specific scientific charts, and interactive or animated formats) remains an open question that falls outside the scope of our current counterfactual framework. Additionally, our counterfactual design manipulates only numerical values. Non-numerical encoding dimensions, including color, spatial layout, and temporal ordering, remain largely unexplored and present a promising avenue that may reveal distinct prioritization mechanisms. Finally, we}{, but may not generalize to more complex visualizations (e.g., network diagrams, 3D plots, or domain-specific scientific visualizations). Additionally, we} restrict our experiments to English prompts, which may overlook language-specific behaviors in multilingual or non-English LVLMs.

\textbf{Methodological Considerations}: The CVLAT suite comprises 48 items, which is sufficient for identifying broad behavioral patterns but constrained for fine-grained or per-task analysis. \replaced{In addition, the VFRI metric should be interpreted in conjunction with the false-response rate. A near-zero VFRI does not, by itself, indicate a balanced integration of sources; intermediate values can also arise from elevated false-response (neither visually nor factually correct) rates, signaling model instability or confusion rather than deliberate arbitration.}{Moreover, our counterfactual design manipulates only numerical values; other dimensions (e.g., color encodings, spatial arrangements, temporal orderings) remain unexplored, and may reveal different forms of visual-factual conflict or different prioritization patterns.}

\deleted{\textbf{Missing Human Baseline}: We do not conduct human experiments with CVLAT. Understanding how humans navigate visual-factual conflicts would provide a critical reference point for interpreting LVLM behavior. While prior work suggests that humans also exhibit knowledge-driven biases \cite{xiong2019curse}, the extent of such biases under counterfactual visualization settings remains unknown.}

\textbf{Future Directions}. Promising directions for future work include the following. \replaced{(1) Investigating whether structural and architectural interventions---such as targeted fine-tuning and data-centric methods~\cite{huang2025evochart, das2025charts, xu2024chartmoe, masry2025chartgemma}, as well as controlled vision-encoder and connector ablations---can achieve cleaner decoupling or more predictable steering of visual--factual prioritization gradients. (2) Systematically varying conflict magnitude (ranging from subtle discrepancies to full reversals) and partial conflicts (where only a specific subset of data points is perturbed) to map precisely how model override rates depend on mismatch strength. (3) Exploring richer prompt designs beyond the binary priority framings evaluated in this work. In particular, developing conflict-aware prompts that explicitly alert the model to potential visual--factual discrepancies, alongside incorporating few-shot demonstrations that contextually anchor visual override behaviors. (4) Resolving the underlying causality behind the F-priority-collapse pattern: while our output-length analysis is consistent with a hypothesis of heightened deliberation-activation, decoupling this from alternative explanations will require controlled, mechanism-level probes (e.g., matched conditional/unconditional instructions and direct access to internal model states). (5) Developing adaptive evaluation frameworks that algorithmically or systematically adjust task difficulty based on an individual model's underlying baseline capabilities, mitigating both floor and ceiling effects across highly disparate model tiers.}{(1) conducting human subject experiments with CVLAT to establish comparative baselines, (2) investigating whether fine-tuning on counterfactual visualizations can better decouple visual interpretation from factual knowledge, (3) developing adaptive evaluation frameworks that adjust task difficulty based on model capabilities, (4) exploring architectural modifications for more controllable information prioritization.}
\section{Conclusion}
This work provides a systematic examination of LVLMs' visualization literacy capabilities and exposes fundamental limitations in how it is currently evaluated. By introducing a quadrant framework that disentangles visual correctness from factual correctness, we show that existing accuracy-based evaluation methods often conflate genuine visualization interpretation with reliance on pre-trained knowledge. As a result, high performance does not necessarily imply faithful visualization understanding. Our experiments with \replaced{15}{12} state-of-the-art LVLMs\added{, complemented by a human baseline study ($N=30$, recruited via Prolific),} demonstrate \replaced{five}{four} main findings. First, several SOTA LVLMs \added{achieve visualization-literacy performance approaching or exceeding human baselines. Under the Normal prompt condition, Gemini-3.1-Pro reaches 99.7\% on VLAT and 93.3\% on reVLAT, while Claude-Opus-4.7 reaches 94.9\% on VLAT and 88.5\% on reVLAT}. Second, \added{the evaluated suite tilts visualization-oriented on average, but a factual knowledge-oriented minority shows statistically significant factual orientation that masks their underlying visualization literacy whenever visual and factual signals conflict.} Third, our proposed CVLAT test reveals distinct model groups—factual knowledge-oriented\deleted{,} and visualization-oriented—with measurable differences in information prioritization. Fourth, prompt engineering can alter information prioritization\added{, but its effectiveness is highly model-dependent and often direction-asymmetric. Factual-priority instruction sometimes shifts models toward visual rather than factual, and one-sided prompt failure occurs in either direction (V- or F-priority-insensitive), indicating that prompt-controllability is often not reciprocal and cannot be assumed from benchmark performance alone}. \added{Fifth, a human baseline shows that humans systematically follow the chart rather than overriding it with factual priors (Sec.~\ref{sec:exp2_results}, Appendix~C).} Overall, these findings have critical implications for deploying LVLMs in visual analytics systems, as practitioners must carefully consider not only overall accuracy, but also how models prioritize visual and factual information. More broadly, our work establishes a foundation for developing more nuanced evaluation frameworks and \deleted{providing practical implications for }understanding the interplay between visual interpretation and factual knowledge in multimodal foundation models.





\bibliographystyle{IEEEtran}
\bibliography{reference}

\newpage
\,
\newpage

 

\begin{IEEEbiography}[{\includegraphics[width=1in,height=1.25in,clip,keepaspectratio]{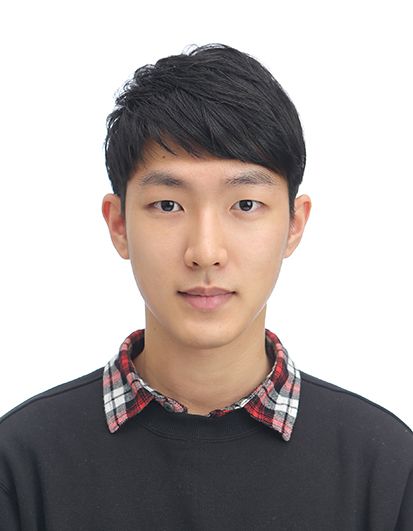}}]{Soohyun Lee}
is a Ph.D. student at the Human-Computer Interaction Laboratory under the Department of Computer Science and Engineering, Seoul National University, Korea. His research interests include Human-AI Interaction, Graphical Perception, and Data Analysis. Before starting his Ph.D. Program, he received a B.S. degree in Computer Science and Engineering and a B.A. degree in Statistics from Korea University.
\end{IEEEbiography}

\begin{IEEEbiography}[{\includegraphics[width=1in,height=1.25in,clip,keepaspectratio]{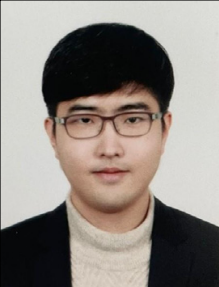}}]{Jaeyoung Kim}
is the Chief Technical Officer at MADI Inc., where he leads the development of AI-driven clinical trial design optimization. His research interests include visual analytics, human-AI interaction, and machine learning for healthcare. He received the B.S. degree in Electrical and Computer Engineering in 2016 from Sungkyunkwan University and the Ph.D. degree in Computer Science and Engineering in 2025 from Seoul National University.
\end{IEEEbiography}

\begin{IEEEbiography}[{\includegraphics[width=1in,height=1.25in,clip,keepaspectratio]{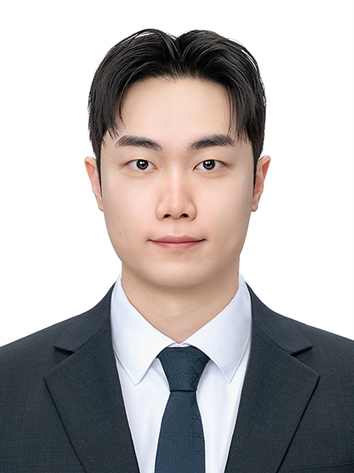}}]{Seokhyeon Park}
is a Postdoctoral Researcher at Seoul National University, Republic of Korea. His research interests include Human-AI Interaction, Visualization, and Interface Design. He received the B.S. degree in Computer Science and Engineering and the B.A. degree in Information Science and Culture in 2020, and the Ph.D. degree in Computer Science and Engineering in 2026, all from Seoul National University.
\end{IEEEbiography}

\begin{IEEEbiography}[{\includegraphics [width=1in,height=1.25in,clip, keepaspectratio]{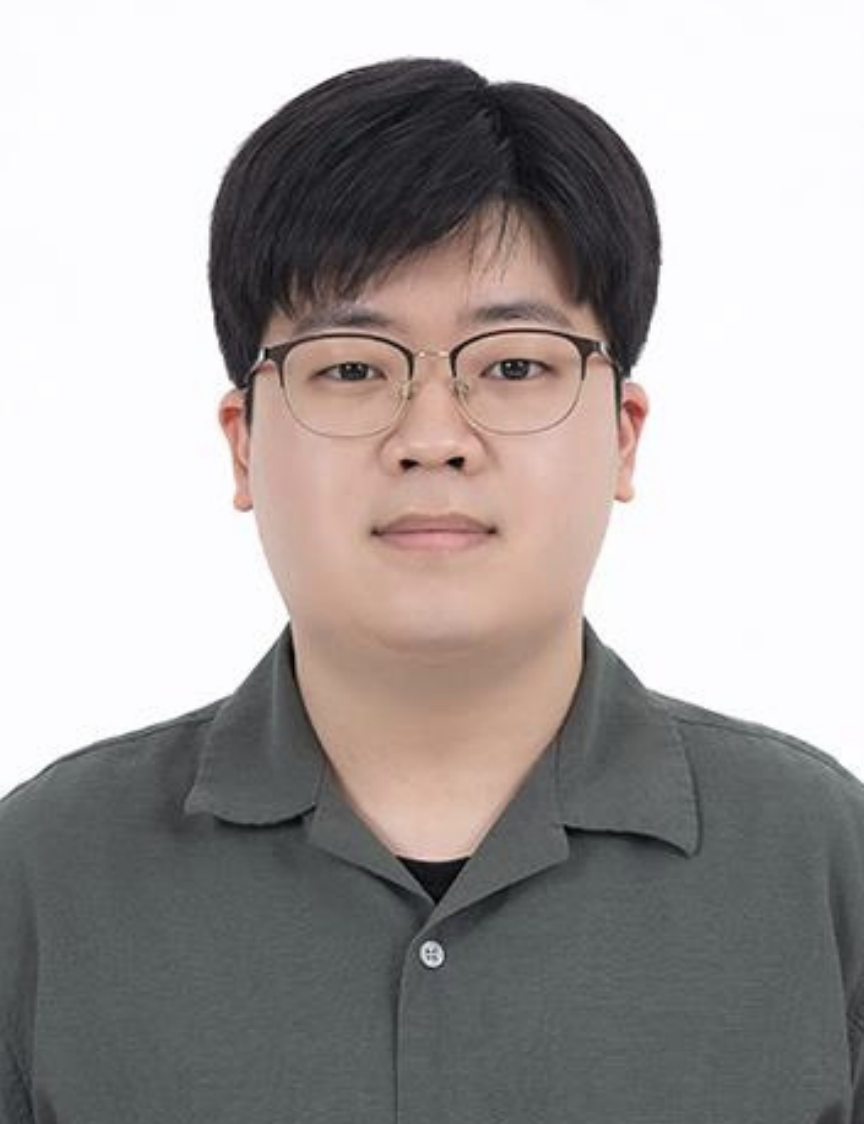}}] {Sihyeon Lee}
is a Ph.D. Student at the Department of Computer Science and Engineering, Seoul National University, Korea. His research interests include Biomedical Informatics, Precision Health, and Medical LLM Agents. Before starting his Ph.D. program, he received a B.S. degree in AI Convergence from Soongsil University.

\end{IEEEbiography}

\begin{IEEEbiography}[{\includegraphics [width=1in,height=1.25in,clip, keepaspectratio]{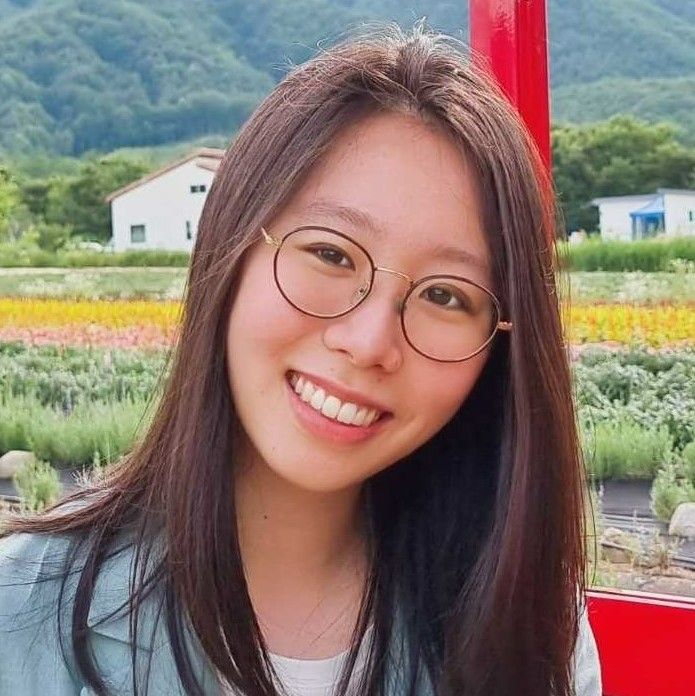}}] {Jiwon Song} 
is a Ph.D. student at the Human-Computer Interaction Laboratory under the Department of Computer Science and Engineering, Seoul National University, Korea. Her research interests are in designing systems that help seniors manage their personal data more effectively. Currently, she is researching a system aimed at enabling seniors to better manage their blood pressure data, making it more accessible and understandable for them.
\end{IEEEbiography}

\begin{IEEEbiography}[{\includegraphics [width=1in,height=1.25in,clip, keepaspectratio]{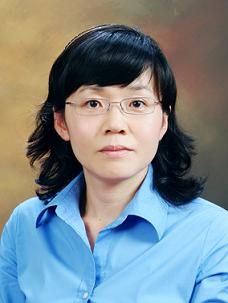}}] {Bohyoung Kim} received the B.S. and M.S. degrees in computer science and the Ph.D. degree in computer science and engineering from Seoul National University, Seoul, South Korea, in 1995, 1997, and 2001, respectively. She is currently a professor in the Department of Biomedical Engineering, Hankuk University of Foreign Studies, South Korea. Her research interests include computer graphics, visualization, medical imaging, and bio-medical informatics.
\end{IEEEbiography}

\begin{IEEEbiography}[{\includegraphics [width=1in,height=1.25in,clip, keepaspectratio]{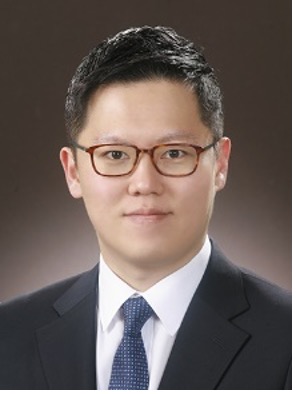}}] {Hyunjoo Song} is currently an assistant professor at the School of Computer Science and Engineering, Soongsil University, South Korea. His research interests include human-computer interaction, information visualization, visual analytics, eye tracking and health informatics. He received the B.S. degree in computer science and engineering in 2009, the M.S. degree in electrical engineering in 2011, and the Ph.D. degree in electrical engineering and computer science in 2016, all from Seoul National University, Seoul, Korea.
\end{IEEEbiography}

\begin{IEEEbiography}[{\includegraphics [width=1in,height=1.25in,clip, keepaspectratio]{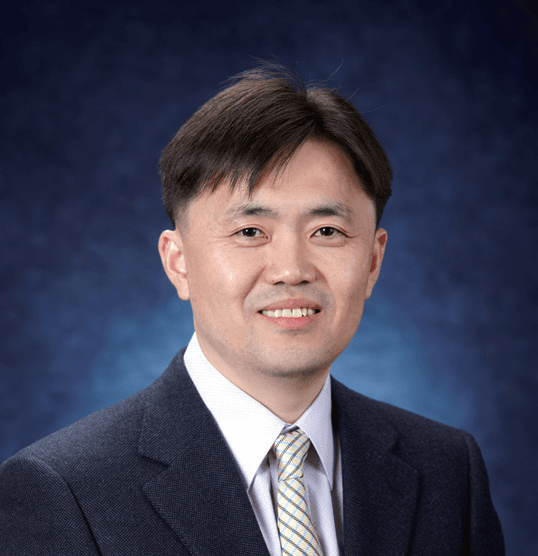}}] {Jinwook Seo} 
is a professor in the Department of Computer Science and Engineering, Seoul National University, where he is also the Director of the Human-Computer Interaction Laboratory. His research interests include Human-Computer Interaction, Information Visualization, and Biomedical Informatics. He received his PhD in Computer Science from the University of Maryland at College Park in 2005.
\end{IEEEbiography}

\vfill
\clearpage
{
\appendices

\section{Prompt Templates}
\label{app:prompts}

This appendix provides the full text of all prompt templates used in our experiments.

\subsection{Experiment 1: VLAT and reVLAT Evaluation}
\label{app:prompts_exp1}

\noindent\textbf{Normal Prompt.} A minimal prompt designed to elicit direct answers without requiring explanation, based on Hong et al.'s \cite{hong2025llms} approach.

\begin{tcolorbox}[colback=gray!5, colframe=gray!40, title=Normal Prompt]
\small
You are a helpful assistant for analyzing data visualizations. Please answer with the letter corresponding to the best option, or choose Omit if unsure. For example, if option (a) is correct, only reply with (a).
\end{tcolorbox}

\noindent\textbf{Explain Prompt.} A chain-of-thought prompt that guides models through structured reasoning steps aligned with Lundgard and Satyanarayan's semantic content framework \cite{lundgard2021accessible}.

\begin{tcolorbox}[colback=gray!5, colframe=gray!40, title=Explain Prompt]
\small
You are a helpful assistant for analyzing data visualizations. When answering the question, please explain your process in this order:\\

1. First, describe the specific points or areas where your attention focused (your eye fixation points) when examining the visualization, and in what order. Be specific about which areas, elements, or coordinates you focused on.\\

2. Next, tell me what specific data values you extracted from each area you examined. Include any numerical values, textual information (labels, legends, categories), units, and trends you observed.\\

3. Then, explain how you used this information to solve the problem, including any calculations, comparisons, or interpretations you made.\\

4. Finally, provide your answer as a single letter (e.g., (a), (b), etc.). If you are unsure, choose Omit.\\

Be as detailed and specific as possible at each step of your analysis.
\end{tcolorbox}

\subsection{Experiment 3: Prompt Engineering for Information Prioritization}
\label{app:prompts_exp3}

\noindent\textbf{Factual Priority Prompt.} Instructs models to prioritize pre-trained factual knowledge over visual information when conflicts arise. Used to test whether visualization-oriented models can be redirected toward factual reliance.

\begin{tcolorbox}[colback=gray!5, colframe=gray!40, title=Factual Priority Prompt]
\small
You are a helpful assistant for analyzing data visualizations. While examining the visualization, prioritize your pre-existing factual knowledge about the real world over what is shown in the visualization if there are contradictions. If the visualization shows information that contradicts well-established facts, rely on your factual knowledge rather than the visual content. Please answer with the letter corresponding to the option that best aligns with factual accuracy, or choose Omit if unsure. For example, if option (a) is correct based on factual knowledge, reply with (a).
\end{tcolorbox}

\noindent\textbf{Visual Priority Prompt.} Instructs models to treat visual data as ground truth and ignore conflicts with factual knowledge. Used to test whether factual knowledge-oriented models can be redirected toward visual fidelity.

\begin{tcolorbox}[colback=gray!5, colframe=gray!40, title=Visual Priority Prompt]
\small
You are a helpful assistant for analyzing data visualizations. Focus exclusively on what is shown in the visualization, even if it contradicts your pre-existing knowledge about the world. Your task is to interpret the visualization exactly as presented, treating the visual data as the ground truth regardless of whether it aligns with real-world facts you may know. Ignore any inconsistencies between the visualization and your factual knowledge—the visualization's context always takes precedence. Please answer with the letter corresponding to the option that best matches what is explicitly shown in the visualization, or choose Omit if unsure. For example, if option (a) is correct based on the visualization data, reply with (a).
\end{tcolorbox}

\subsection{\added{Capability-Bound Conditions}}\label{app:probe_prompts}

\added{The CVLAT protocol additionally administers two control conditions that supply the capability references entering VF and FA (Sec.~V).}

\noindent\added{\textbf{Anonymized visual baseline.} Same counterfactual chart as CVLAT, but axis labels and category names are replaced with neutral placeholders so the model has no factual signal to recall. Estimates chart-reading capability $V_{\text{anon}}$.}

\noindent\added{\textbf{Q-only condition.} The same domain question with the chart removed. Estimates factual-prior availability $F_Q$.}

\begin{tcolorbox}[colback=gray!5, colframe=gray!40, title=Q-only Probe]
\small
\added{You will be given a question and a set of choices. Use your pre-existing factual knowledge about the real world to choose the best option. Please answer with the letter corresponding to the best option, or choose Omit if you do not know. For example, if option (a) is correct, reply with (a).}
\end{tcolorbox}

\section{Detailed Performance Analysis Across Chart and Task Types}
\label{app:exp1_performance}

This appendix presents a comprehensive performance analysis of all evaluated LVLMs across different chart types and task types, comparing both VLAT and reVLAT benchmarks under normal and explain prompt conditions.

To provide clearer insights into model capabilities, we consolidated task types by removing parenthetical subtypes. The consolidation mapping is as follows:
\begin{itemize}
    \item Retrieve Value (absolute value, relative value, derived value) $\rightarrow$ Retrieve Value
    \item Make Comparisons (absolute value, relative value, derived value) $\rightarrow$ Make Comparisons
    \item Find Extremum (relative value, derived value) $\rightarrow$ Find Extremum
    \item All other task types remain unchanged
\end{itemize}

This consolidation allows for more meaningful comparisons across models and reduces noise from overly specific task categorizations.

\noindent\textbf{Performance by Chart Type.} Figure~\ref{fig:chart_type_performance} shows accuracy across 12 chart types. \replaced{The top-performing frontier models (Gemini-3.1-Pro and Claude-Opus-4.7) consistently outperform}{Gemini-2.5-Pro consistently outperforms} other models across most chart types, while models generally struggle with treemaps and stacked bar charts. The heatmap reveals that performance gaps between VLAT and reVLAT vary substantially by chart type, with some visualizations (e.g., pie charts and treemaps) showing larger drops than others.

\begin{figure*}
\centering
\includegraphics[width=\textwidth]{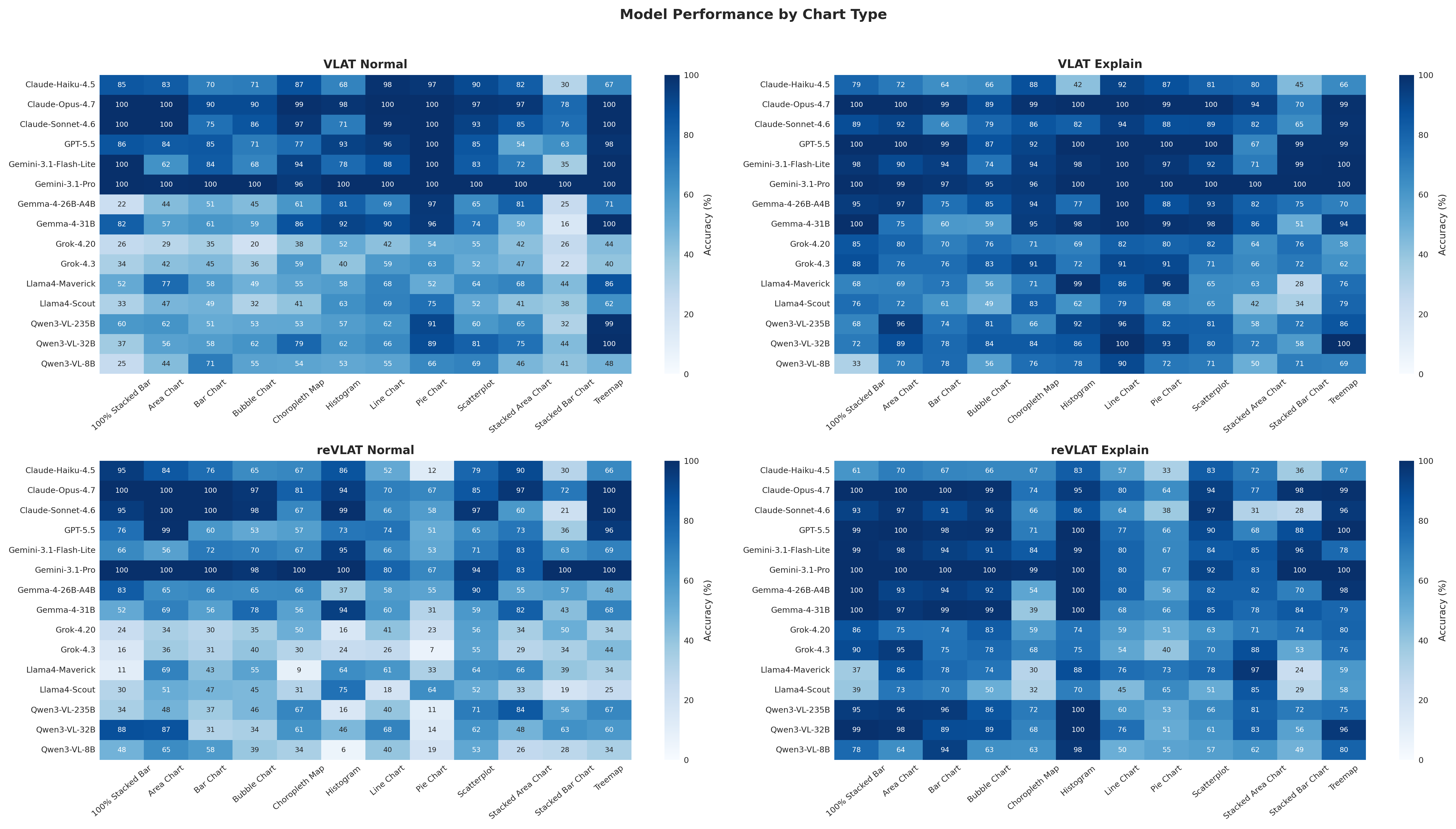}
\caption{Model performance across chart types in Experiment 1. Accuracy (\%) for VLAT and reVLAT benchmarks with normal and explain prompts.}
\label{fig:chart_type_performance}
\end{figure*}

\noindent\textbf{Performance by Task Type.} Figure~\ref{fig:task_type_performance} presents accuracy across 8 consolidated task categories. Models show relatively strong performance on \textit{\replaced{Identify}{Indentify} Hierarchical Structure} and \textit{\replaced{Find}{Fine} Clusters} tasks, while \textit{Find Anomalies} and \textit{Find Correlations/Trends} (in VLAT) tasks prove more challenging.

\begin{figure*}
\centering
\includegraphics[width=\textwidth]{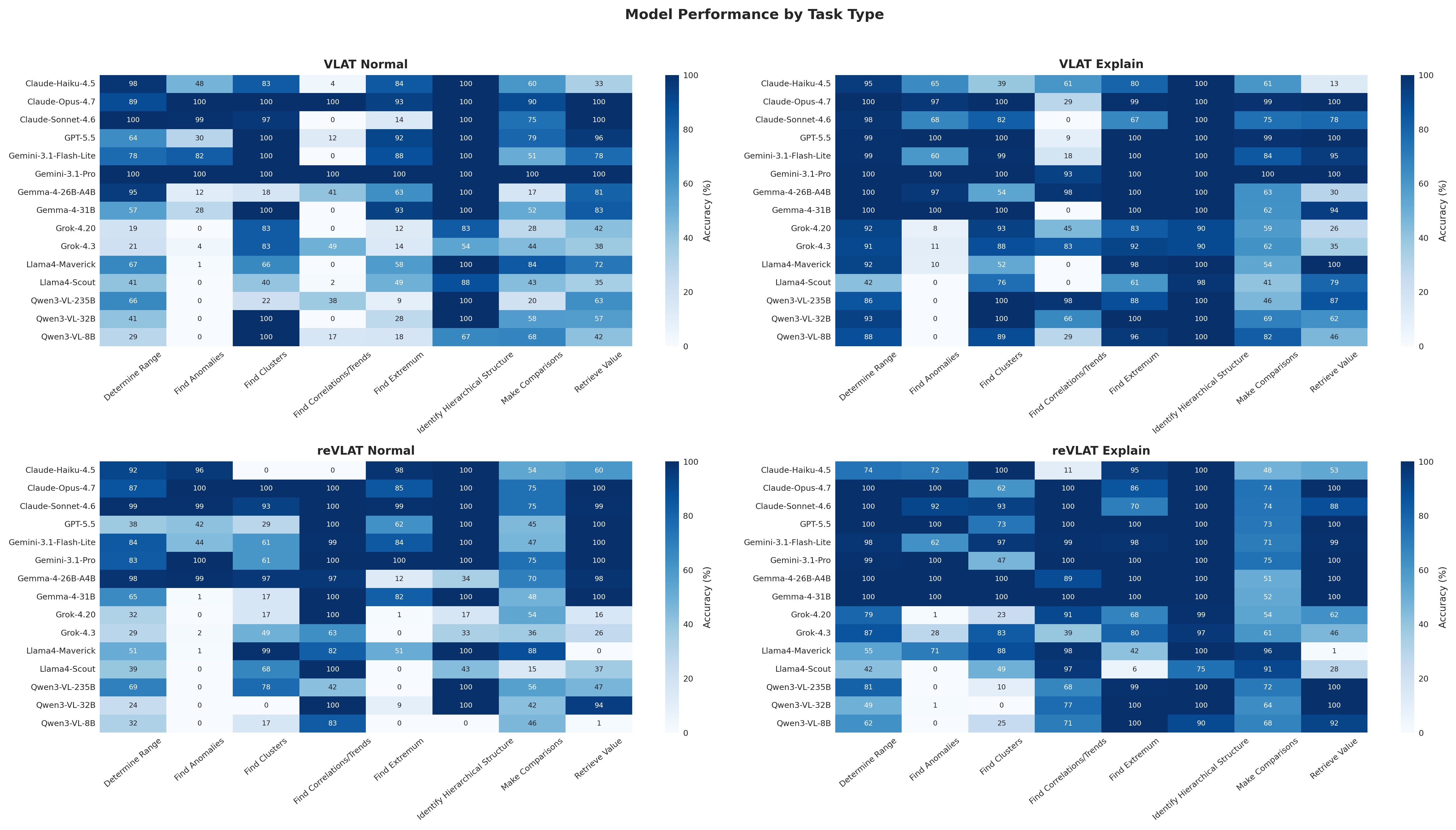}
\caption{Model performance across consolidated task types (8 categories) in Experiment 1 for VLAT and reVLAT benchmarks.}
\label{fig:task_type_performance}
\end{figure*}

\noindent\textbf{VLAT vs. reVLAT Comparison.} Figure~\ref{fig:vlat_revlat_comparison} directly compares performance between VLAT and reVLAT across chart types (left) and task types (right). The stacked bars illustrate the additive effect of the Explain prompt. This comparison highlights where factual-knowledge confounding is most pronounced: larger gaps between VLAT and reVLAT are consistent with stronger reliance on pre-trained factual priors, in line with reVLAT's motivation that aligned VLAT items may be solvable from learned real-world knowledge rather than visual interpretation alone.

\begin{figure*}
\centering
\includegraphics[width=\textwidth]{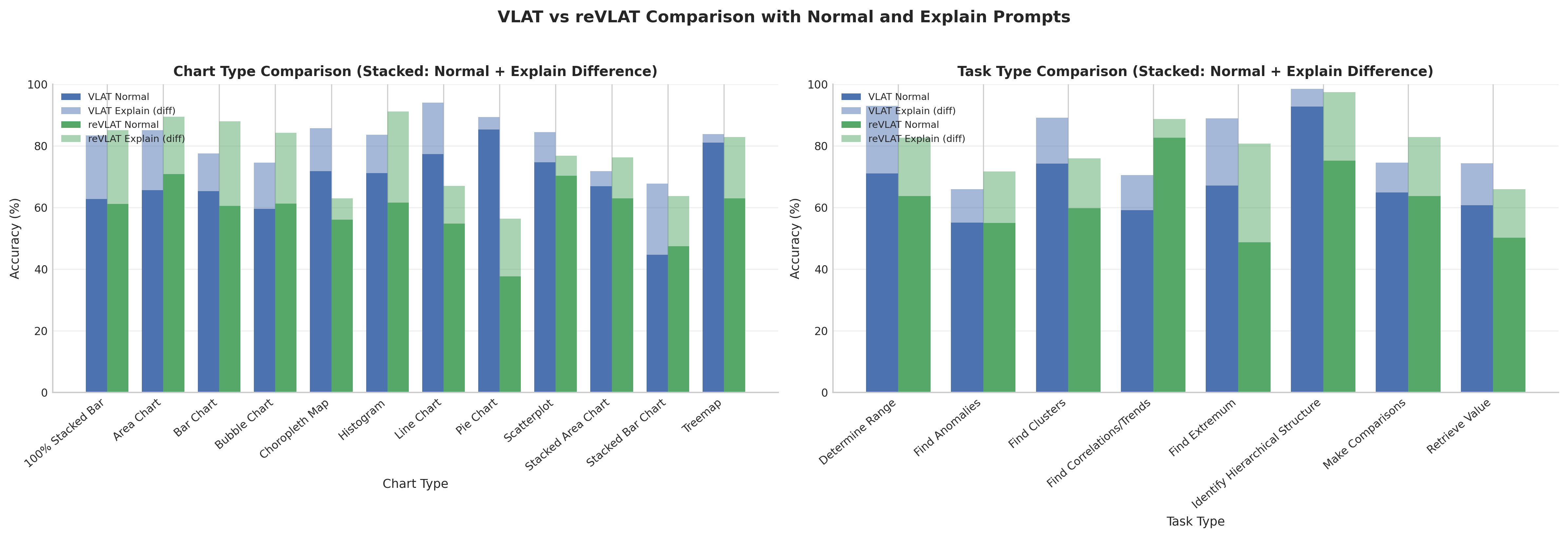}
\caption{VLAT vs reVLAT comparison in Experiment 1. Left: chart type performance. Right: task type performance. Stacked bars show normal prompt accuracy (solid) with explain prompt difference (transparent).}
\label{fig:vlat_revlat_comparison}
\end{figure*}

\section{\added{Human Study Protocol}}\label{app:human_protocol}

\added{This appendix documents the CVLAT human baseline study supporting Sec.~V-A2 (design) and Sec.~V-B (results).}

\added{\textbf{Ethics.} The study was reviewed and approved by the authors' institutional review board (IRB).}

\added{\textbf{Recruitment.} $N = 30$ participants recruited via Prolific with the same screening criteria as a recent Prolific VLAT replication~\cite{pandey2023mini}: nationality and current country of residence both restricted to the United States, English as first language and fluent language, approval rate 95--100\%, and 1{,}000--100{,}000 prior submissions.}

\added{\textbf{Materials.} The same 48 CVLAT items used in the LVLM evaluation, presented with the same multiple-choice options (visually-correct, factually-correct, distractor(s), Omit).}

\added{\textbf{Procedure.} Each participant completed all 48 items (one per page) with option order randomized per participant. Participants were instructed that the questions concern visualization interpretation, without disclosure of the counterfactual nature of the data.}

\added{\textbf{Scoring.} Each participant's score follows the original VLAT correction-for-guessing convention: $CS = R - W/(C-1)$, where $R$ is the number of target-correct responses, $W$ the number of incorrect responses (i.e., wrong-option selections, excluding Omits and timeouts), and $C$ the number of options per item. Omit and timeout responses incur no penalty. We use the standard $C-1$ correction here for comparability with the published VLAT human baseline, which applies the same convention. The LVLM VF/FA normalization (and the per-participant human VFRI reported below) instead uses $C_i - 2$, because both the visually-correct and factually-correct options are focal response categories rather than distractors, leaving $C_i - 2$ true distractors against which to correct for guessing.}

\added{\textbf{Results (Visual as target).} Mean corrected accuracy is $53.71\%$ (SD $14.92$), and mean raw accuracy is $60.76\%$ (SD $11.70$). The corrected score is statistically equivalent (within a $\pm 10$ pp margin; see the equivalence test below) to the original VLAT human baseline ($51.91\%$, SD $16.57$, $N=191$~\cite{lee2016vlat}) and to a recent Prolific VLAT replication ($53.08\%$, SD $18.96$, $N=199$~\cite{pandey2023mini}), supporting that CVLAT items are not systematically harder than VLAT items for humans (Sec.~V-A2).}

\added{\textbf{Statistical equivalence test.} To formally support the calibration claim, we conducted a two one-sided test (TOST) with a pre-specified equivalence margin of $\pm 10$ pp. Both comparisons reject the non-equivalence hypothesis at $\alpha = 0.05$: against Lee et al.~\cite{lee2016vlat}, $p=0.003$ with a 90\% CI on the difference of $[-3.10, +6.70]$ pp; against the Mini-VLAT replication~\cite{pandey2023mini}, $p=0.001$ with 90\% CI $[-4.37, +5.63]$ pp. Both intervals are contained within $\pm 10$ pp, supporting equivalence at this margin (we do not claim equivalence at any tighter margin).}

\added{\textbf{Results (Factual as target).} Mean corrected accuracy is $-9.46\%$ (SD $8.97$), and mean raw accuracy (factual-correct rate) is $13.26\%$ (SD $7.69$). For this diagnostic we report the \emph{unclamped} correction-for-guessing score $R - W/(C-1)$ (without the $\max(0, \cdot)$ floor that Equation~1 applies for LVLM VF/FA normalization). The corrected score is well below chance level, indicating that participants in aggregate did not pick the factually-correct option that contradicted the chart.}

\added{\textbf{Per-participant VFRI.} Applying the per-question VFRI formula (Eq.~4, with the same $C_i - 2$ denominator used for LVLMs) to each participant's 48 chart responses yields a participant-level VFRI distribution with mean $+0.64$ (SD $0.21$, range $[+0.12, +0.94]$). All 30 participants score positive VFRI, i.e., every participant is visualization-oriented. Note that we did not administer the anonymized-baseline or Q-only conditions to human participants, so these per-participant VFRI values are computed without capability normalization.}

\section{\added{Model Snapshots and Inference Configuration}}\label{app:model_snapshot}

\added{This appendix records the exact model identifiers and inference configurations used in our evaluation, enabling reproducibility. API models accessed via OpenRouter are reported with the model snapshot used in our API calls (Table~\ref{tab:model_api}). Locally hosted models are reported with their Hugging Face checkpoint and quantization (Table~\ref{tab:model_local}).}

\begin{table*}[!t]
\centering
\caption{\added{API-accessed models via OpenRouter. We report the exact model snapshot and reasoning mode used in our API calls. Reasoning was minimized at each endpoint: disabled where supported by the API, set to the lowest available value otherwise. For proprietary API models, parameter counts and architecture are not publicly disclosed; Qwen3-VL-235B was accessed via OpenRouter due to its scale, while the remaining open-source checkpoints (Table~\ref{tab:model_local}) were hosted locally.}}
\label{tab:model_api}
\small
\begin{tabular}{llll}
\toprule
\textbf{Model} & \textbf{Provider} & \textbf{Model snapshot used} & \textbf{Reasoning mode} \\
\midrule
Claude-Haiku-4.5      & Anthropic & \texttt{anthropic/claude-4.5-haiku-20251001} & off \\
Claude-Sonnet-4.6     & Anthropic & \texttt{anthropic/claude-4.6-sonnet-20260217} & off \\
Claude-Opus-4.7       & Anthropic & \texttt{anthropic/claude-4.7-opus-20260416} & off \\
Gemini-3.1-Flash-Lite & Google    & \texttt{google/gemini-3.1-flash-lite-preview-20260303} & minimal \\
Gemini-3.1-Pro        & Google    & \texttt{google/gemini-3.1-pro-preview-20260219} & low \\
GPT-5.5               & OpenAI    & \texttt{openai/gpt-5.5-20260423} & none \\
Grok-4.20             & xAI       & \texttt{x-ai/grok-4.20-20260309} & off \\
Grok-4.3              & xAI       & \texttt{x-ai/grok-4.3-20260430} & none \\
Qwen3-VL-235B         & Alibaba   & \texttt{qwen/qwen3-vl-235b-a22b-instruct}\,$^{\dagger}$ & n/a \\
\bottomrule
\end{tabular}

\vspace{2pt}
\footnotesize $^{\dagger}$\,OpenRouter does not expose a dated suffix for this snapshot.
\end{table*}

\begin{table*}[!t]
\centering
\caption{\added{Locally hosted models. All local models were served with vLLM 0.20.1 on on-prem servers with NVIDIA RTX A5000 GPUs.}}
\label{tab:model_local}
\footnotesize
\setlength{\tabcolsep}{4pt}
\begin{tabular}{l l >{\raggedright\arraybackslash}p{2.6in} l l l}
\toprule
\textbf{Model} & \textbf{Provider} & \textbf{HF checkpoint} & \textbf{Quant.} & \textbf{Params} & \textbf{Arch.} \\
\midrule
Qwen3-VL-8B     & Alibaba & \url{cyankiwi/Qwen3-VL-8B-Instruct-AWQ-4bit}                       & AWQ 4-bit & 8B                       & dense \\
Qwen3-VL-32B    & Alibaba & \url{cyankiwi/Qwen3-VL-32B-Instruct-AWQ-4bit}                      & AWQ 4-bit & 33B                      & dense \\
Gemma-4-26B-A4B & Google  & \url{cyankiwi/gemma-4-26B-A4B-it-AWQ-4bit}                         & AWQ 4-bit & 25.2B / 3.8B$^{*}$      & MoE \\
Gemma-4-31B     & Google  & \url{cyankiwi/gemma-4-31B-it-AWQ-4bit}                             & AWQ 4-bit & 30.7B                    & dense \\
Llama4-Maverick & Meta    & \url{RedHatAI/Llama-4-Maverick-17B-128E-Instruct-quantized.w4a16}  & W4A16     & $\sim$400B / 17B$^{*}$  & MoE \\
Llama4-Scout    & Meta    & \url{RedHatAI/Llama-4-Scout-17B-16E-Instruct-quantized.w4a16}      & W4A16     & $\sim$109B / 17B$^{*}$  & MoE \\
\bottomrule
\end{tabular}

\vspace{2pt}
\footnotesize $^{*}$\,Active parameters per token. Gemma-4-26B-A4B activates 8 of 128 experts, Llama4-Maverick has 128 experts, and Llama4-Scout has 16 experts.
\end{table*}
}
\end{document}